\documentclass[journal]{IEEEtran}
\usepackage{cite}
\usepackage{amsmath,amssymb}

\usepackage{subfig}
\usepackage{graphicx}
\usepackage{multirow}

\usepackage{xcolor}
\usepackage{wrapfig}
\usepackage{algorithm}
\usepackage{algcompatible}
\usepackage{caption}

\usepackage[colorlinks=false]{hyperref}
\hypersetup{hidelinks}
\usepackage{fancyhdr}

\fancypagestyle{firstpage}
{
    \fancyhead[L]{}    
    \fancyhead[R]{\textit{Published at IEEE Transactions on Neural Networks and Learning Systems, DOI: 10.1109/TNNLS.2023.3261860}}
}

\begin{document}

\title{Central-Smoothing Hypergraph Neural Networks for Predicting Drug-Drug Interactions}

\author{Duc~Anh~Nguyen,
Canh~Hao~Nguyen, and
Hiroshi~Mamitsuka \IEEEmembership{Senior Member,~IEEE}

\thanks{The authors are with Bioinformatics Center, Institute for Chemical Research, Kyoto University, Japan. H. M. is also with Department of Computer Science, Aalto University, Finland.} 
\thanks{D.A.N, C.H.N. and H.M. have been supported in part by Otsuka Toshimi Scholarship Foundation, MEXT KAKENHI [%grant number 
22K12150]
 and %in part by %JST ACCEL [grant number JPMJAC1503], 
MEXT KAKENHI [16H02868, 19H04169, 21H05027, 22H03645] 
%FiDiPro by Tekes (currently Business Finland)
(and AIPSE program by Academy of Finland), respectively.}
\thanks{Corresponding author: D.A.N. (email: ducanh@kuicr.kyoto-u.ac.jp).}

\thanks{Published at IEEE Transactions on Neural Networks and Learning Systems, DOI: 10.1109/TNNLS.2023.3261860.}}

\maketitle

\begin{abstract}
Predicting drug-drug interactions (DDI) is the problem of predicting side effects (unwanted outcomes) of a pair of drugs  using drug information and known side effects of many pairs. This problem can be formulated as predicting labels (i.e. side effects) for each pair of nodes in a DDI graph, of which nodes are drugs and edges are interacting drugs with known labels. State-of-the-art methods for this problem are graph neural networks (GNNs), which leverage neighborhood information in the graph to learn node representations. For DDI, however, there are many labels with complicated relationships due to the nature of side effects. Usual GNNs often fix labels as an one-hot vector that does not reflect label relationships and potentially do not obtain the highest performance in the difficult cases of infrequent labels. In this paper, we formulate DDI as a hypergraph where each hyperedge is a triple: two nodes for drugs and one node for a label. We then present $\mathrm{CentSmoothie}$, a hypergraph neural network that \emph{learns representations of nodes and labels altogether} with a novel \emph{'central-smoothing'} formulation. We empirically demonstrate the performance advantages of $\mathrm{CentSmoothie}$ in simulations as well as real datasets.  
\end{abstract}

\section{Introduction} 
\begin{figure}[tp]
     \centering
     \subfloat[][]{\includegraphics[width=0.11\textwidth]{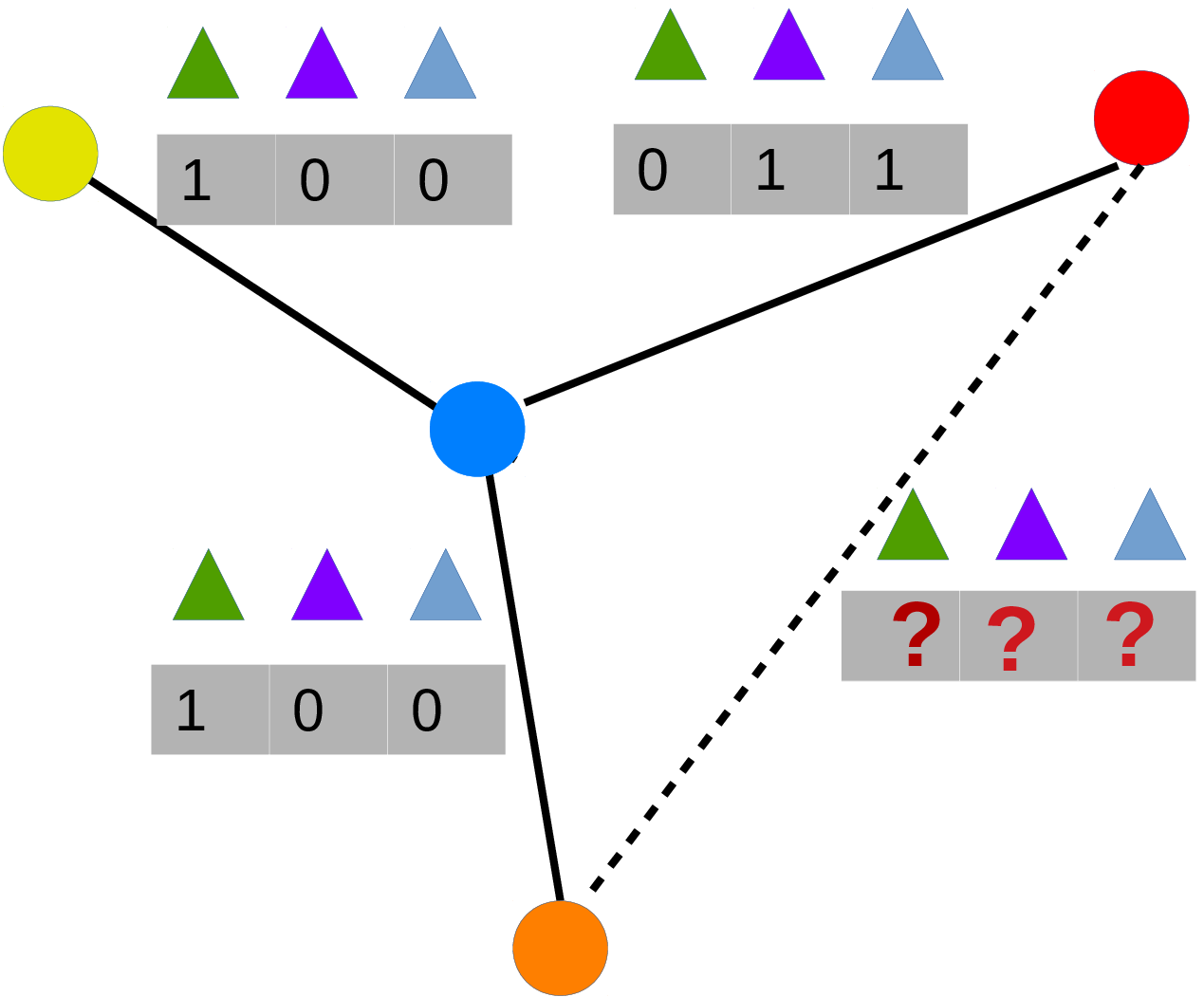}\label{fig:graph}} \hspace{5pt}
     \subfloat[][]{\includegraphics[width=0.11\textwidth]{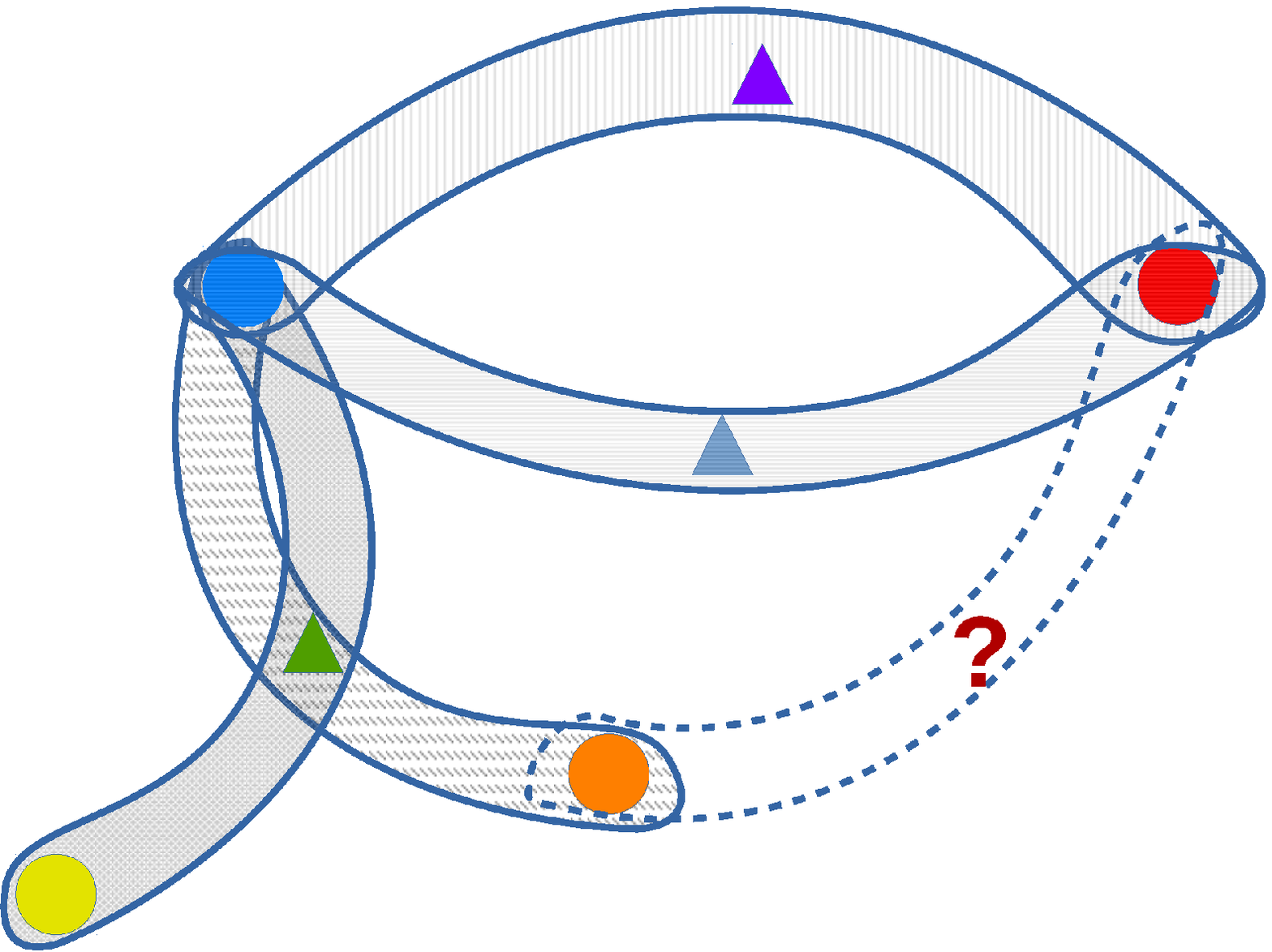}\label{fig:hypergraph}} \hspace{5pt}
     \subfloat[][]{\includegraphics[width=0.22\textwidth]{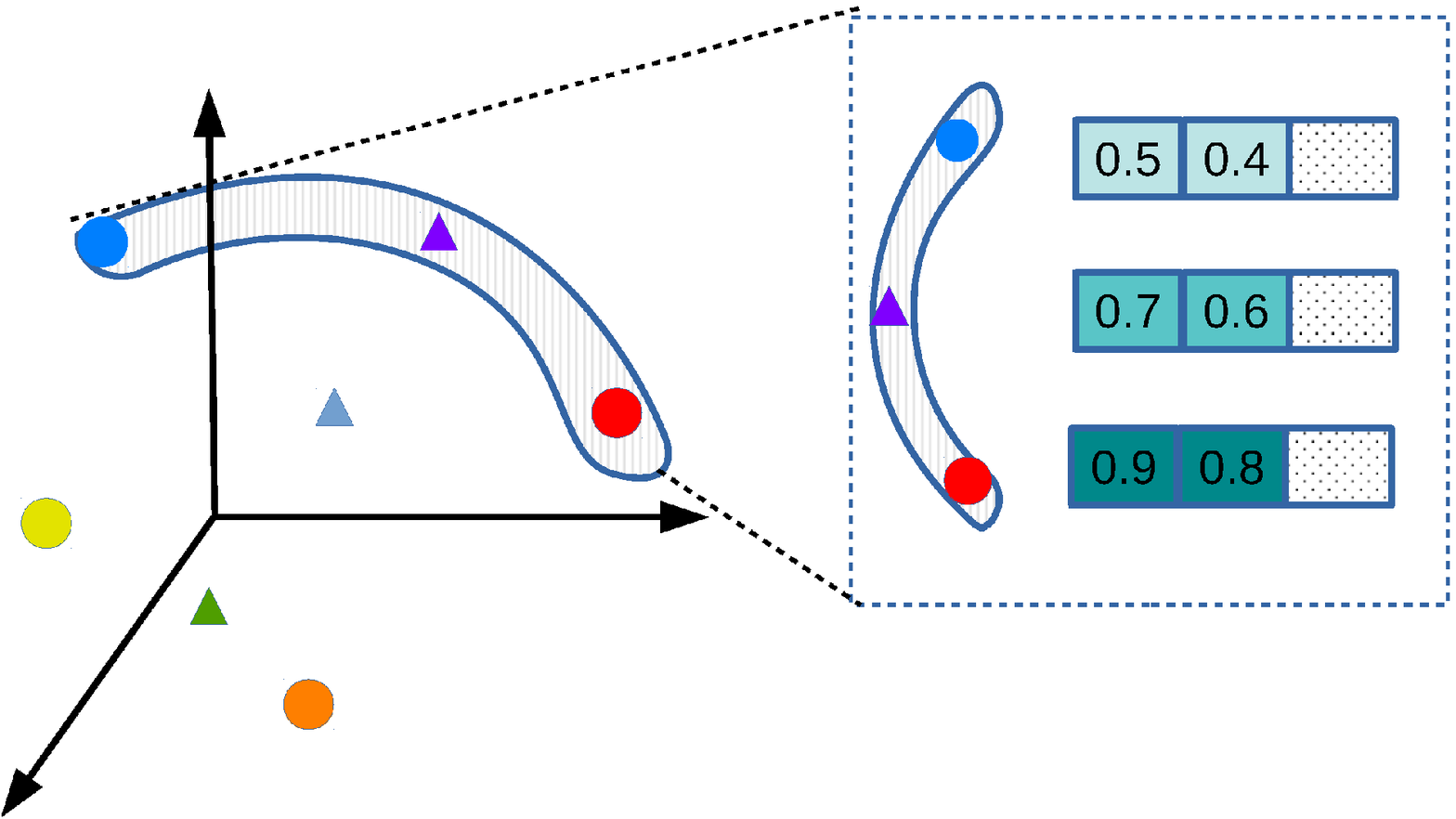}\label{fig:centlatent}} \vspace{-2mm} \ \\
      \subfloat{\includegraphics[width=0.8\linewidth]{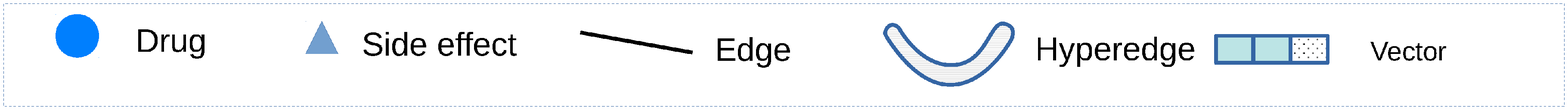}}
          
\caption{Illustrative examples of (a) a traditional graph and (b) a (proposed) hypergraph for drug-drug interactions, and (c) central-smoothing assumption.}
\label{fig:edgevshyper}
\vspace{-2mm}
\end{figure}

In drug-drug interactions (DDI), concurrent use of two drugs can lead to side effects, which are unwanted reactions in human bodies.
It is a very important task to predict drug-drug interactions to guide drug safety. Given drug information and known side effects of many pairs of drugs, one wishes to learn a model to predict side effects of all pairs of drugs, which include new pairs of drugs without known side effects or known pairs (to denoise or complete side effect data). DDI is usually represented as a graph with nodes for drugs, edges for drug pairs that interact, with (binary vector) labels for (known) side effects \cite{zitnik2018modeling}. The task is to predict labels of all pairs of nodes in the DDI graph. Fig. \ref{fig:graph} shows an example of a DDI graph, where the dotted edge with question marks is the pair of drugs with labels to be predicted.

Recently, graph neural networks have emerged as a prominent approach for this task with high prediction performance \cite{zitnik2018modeling,feng2020dpddi}. Graph neural networks for predicting DDI have two steps: learning new representations of drugs from a DDI graph, and using these representations for predictions. One drawback of this approach is the lack of learning label (i.e. side effect) representations. There are many side effects with complicated relationships. For example, our largest dataset has 964 side effects, where the number of drug pairs for one side effect (positive samples in supervised learning) ranges from 288 to 22,520. Previous methods represent each side effect as an independent one-hot vector, potentially under-utilizing the relationship among side effects \cite{zitnik2018modeling,chu2019mlrda,feng2020dpddi}. Considering the relationship between side effects would be beneficial for predicting side effects, especially the ones with only small numbers of positive samples (i.e. infrequent side effects). Hence, it is desirable to learn the representations for both drugs and side effects, namely both nodes and edge labels, together.

To this end, we propose to encode DDI data with a hypergraph \cite{nguyen2020learning}. A node in the hypergraph can be either a drug or a side effect. A hyperedge is a triple of two drugs and a side effect that they caused. Hence, a pair of drugs with multiple side effects will result in many hyperedges in the hypergraph. Fig. \ref{fig:hypergraph}  illustrates an example of a hypergraph corresponding to the DDI graph in Fig. \ref{fig:graph}. 
 Existing learning methods of hypergraph neural networks are based on a \emph{smoothing assumption} that the representations of nodes in a hyperedge should be close to each other \cite{feng2019hypergraph,bai2019hypergraph,yadati2019hypergcn}. However, this assumption is not necessarily appropriate for our DDI problem, since each node representation should reflect (chemical or biological) properties of the corresponding drug and interacting drugs do not necessarily need to have similar properties.
 
 %Figure \ref{fig:smoothing1} illustrates the node representations in a hyperedge of two drugs and one side effect under the traditional smoothing assumption. These representations tend to be close to each other. However, the problem is that two drugs with different properties still interact with each other, hence their representations should not be close.
 
%  \begin{figure}
%      \centering
%      \subfloat[][]{\includegraphics[width=0.14\linewidth]{figs/CentralA1.eps}\label{fig:smoothing1}}
%      \subfloat[][]{\includegraphics[width=0.38\linewidth]{figs/CentralA2.eps}\label{fig:smoothing2}}\\
     
%          \subfloat{\includegraphics[width=0.4\linewidth]{figs/LegendA.eps}}
           
%   \caption{Illustrative examples of two hyperedges. For a hyperedge with two drug nodes and one side effect node, (a) traditional smoothing assumption where all nodes in a hyperedge should be close to each other and (b) the central-smoothing assumption where the side effect node should be close to the midpoint (center) of the two drugs.}
% \end{figure}
 
We propose $\mathrm{CentSmoothie}$, a central-smoothing hypergraph neural network that uses our idea, \emph{central-smoothing assumption} (see Fig. \ref{fig:centlatent}) for each hyperedge in the hypergraph for DDI.  The idea is to learn node representations in a hyperedge such that (i) a drug node representation reflects the property of the corresponding drug and (ii) a side effect node representation reflects a combination of some properties of the two drugs that cause the corresponding side effect \cite{leucuta2006pharmacokinetics,corrie2011mechanisms}. To implement (ii), we first assume that a side effect representation should be related to the midpoint of the representations of the two interacting drugs, reflecting the combination of the two drug properties. Furthermore, there might have different side effects of the same two drugs, suggesting that each side effect might be obtained by a partial combination of the two drug properties.  Hence, we propose that the representation for each side effect is learned to be close to a weighted midpoint of the corresponding two drug representations. 

We formulate the above %central-smoothing 
assumption, and then define the central-smoothing hypergraph Laplacian to be used in each layer of the hypergraph neural network with spectral convolution \cite{feng2019hypergraph}. We also provide a computational method with the complexity of O(n) for the proposed  hypergraph Laplacian. 
 
 We conducted extensive experiments to verify the performance advantages of $\mathrm{CentSmoothie}$ in both synthetic and real datasets. Our experimental results demonstrated that $\mathrm{CentSmoothie}$ significantly outperformed  existing spectral-based convolutional hypergraph neural networks in all cases. In particular,  
 $\mathrm{CentSmoothie}$ achieved higher performances over baselines for real datasets with more infrequent side effects, which are more difficult to predict, justifying the benefit of learning label (side effect) representations. 
 \iffalse
 The organization of the paper is as follows. Section 2 presents related work on predicting DDI. Section 3 is for background on hypergraph Laplacian derived from smoothing measures. The proposed method of central-smoothing hypergraph neural networks is in Section~4. The experimental part is in Section~5, and the conclusion is in Section 6.
 \fi

\section{Related Work}

Existing work in predicting DDI can be divided into two approaches: non-graph based and graph based ones. In the non-graph based approach, pre-defined feature vectors, indicating the existences of chemical substructures and interacting proteins of drugs, are used. The  side effects can be predicted by using a model (for example, a multilayer feedforward neural network), which receives the feature vectors of two drugs as input and the vector indicating the side effects of the two drugs as output \cite{rohani2019drug,chu2019mlrda}. 

In the graph based approach, topological information of graphs is used to enhance the representations of nodes, leading to higher performance than the non-graph based approach. There are two types of graphs that can be used: molecular graphs of drugs and a DDI graph. 
%With the molecular graphs of drugs, each drug is a graph of atoms, hence graph neural networks can be applied to learn a new representation vector for the drug \cite{xu2019mr,deac2019drug}. 
For a DDI graph where nodes are drugs and edges are interactions between drugs, graph neural networks (GNNs) are applied to learn a new representation of a drug node based on its neighbors. Recent results show that GNNs for predicting DDI achieves the cutting-edge performance \cite{zitnik2018modeling,feng2020dpddi}. An extension of a DDI graph can be a DDI heterogeneous graph, where nodes are drugs and side effects and edges are pairs of interacting drugs or drug-side effects \cite{zhang2019heterogeneous}. However, the DDI heterogeneous graph cannot preserve triples of drug-drug-side effects. 

GNNs can be further divided into two approaches: spectral convolution and spatial convolution \cite{wu2020comprehensive}. In the spectral convolution, at first, the graph Laplacian is defined, and then each GNN layer is constructed from the graph Fourier transformation given the graph Laplacian  \cite{feng2019hypergraph,kipf2016semi}. The spatial convolution approach uses node spatial relation that a node is updated based on information from neighbor nodes \cite{gilmer2017neural,zhang2019heterogeneous}. 

Different from existing work for predicting drug-drug interactions, we formulate the drug-drug interactions in the form of a hypergraph and develop a new hypergraph neural network (HGNN) on the DDI hypergraph.

In HGNNs, recent work has inherited the spectral convolution approach on graphs to adapt to hypergraphs by defining the hypergraph Laplacian \cite{feng2019hypergraph,yadati2019hypergcn}. Once the hypergraph Laplacian is defined, HGNNs can be constructed in the same manner as that for GNNs. Another approach for HGNNs is the spatial convolution approach with attention mechanisms \cite{bai2019hypergraph}.

\section{Background}

In this section we briefly describe the hypergraph Laplacian being derived from a smoothness measure \cite{nguyen2020learning}. Let $G = (V, E)$ be a general hypergraph, where $V$ is the node set and $E \subset 2^V$ is the hyperedge set. Let $W = diag(w(e_1), ..., w(e_{|E|})) \in \mathbb{R}^{|E| \times |E|} \succcurlyeq \mathbf{0}$ be the diagonal matrix that $w(e)$ is the weight of hyperedge $e$. Let $x \in  R^{|V|}$ be values of nodes on the  hypergraph that $x_u$ is the value of $x$ at node $u$. 

The hypergraph Laplacian is usually defined to be used in a similar manner to the graph Laplacian: to evaluate the smoothness of a function on a graph.
Let $sh(x,G)$ be a smoothness measure of $x$ on $G$ and $ss(x, e)$ be a smoothness measure of $x$ on hyperedge $e$. The smoothness on the hypergraph usually has the following form \cite{nguyen2020learning}:
\begin{align}
sh(x,G) = \mathcal{T}_{e \in E} w(e)ss(x, e)
\end{align}
where $\mathcal{T}$ is an aggregation operator, such as sum (the most commonly used one), max, or $l_p$ norm \cite{nguyen2020learning}. Usual smoothing assumption on hypergraphs is that nodes within a hyperedge should be close to each other \cite{feng2019hypergraph,bai2019hypergraph,chan2020generalizing}, and then the smoothness measure on each hyperedge is calculated by:
\begin{align}
ss(x, e) = \sum_{(u,v) \in e} (x_u - x_v)^2.
\end{align}
When $\mathcal{T}$ is a sum operator, the smoothness of a function on a hypergraph can be found in the following form:
\begin{align}
sh(x,G) &= \sum_{e \in E} w(e) \sum_{(u,v) \in e} (x_u - x_v)^2 \\
        &= x^{\mathbf{T}}Lx,
\end{align}
which has the quadratic form with $L$, and $L$ is then called the hypergraph Laplacian of the hypergraph. In the next section, we will propose a new smoothing assumption on hypergraphs then define a new hypergraph Laplacian.

\section{CentSmoothie: Central-Smoothing Hypergraph Neural Networks}
\iffalse
We formulate a neural network on hypergraph with a central-smoothing assumption for the DDI problem. We first show the problem setting, and then present the central-smoothing hypergraph Laplacian. Once we obtain the hypergraph Laplacian, we can define each layer of the network\cite{feng2019hypergraph}. 
\fi
\subsection{Problem Setting}
We formulate the problem of predicting DDI as follows.

\textit{Input}: Given a hypergraph of drug-drug interactions: $G = (V, E)$, where the node set $V = V_D \cup V_S$ consists of a drug node set $V_D$ and a side effect node set $V_S$, a known hyperedge set $E \subset V_D \times V_D \times V_S$ (Since two drugs in a drug pair are unordered, two triples $(u, v, t) $  and $(v, u, t)$ ($u,v \in V_D$ and $t \in V_S$) are the same), and the feature vectors of drugs: $X_D \in R^{|V_D| \times K_0}$, where $K_0$ is the feature size. The feature vectors of side effects are one-hot vectors.

\textit{Output}: For each triple $e=(u, v, t) \in V_D \times V_D \times V_S$, $t$ is predicted to be a side effect of $u$ and $v$ if the score of the triple is larger than a threshold. 

\subsection{Central-Smoothing Hypergraph Laplacian}

%The key idea is a central-smoothing assumption: each hyperedge is called {\it central-smooth} if a weighted version of the midpoint of drug node representations is close enough to the representation of the side effect node. The midpoint is supposed to contain all combined properties of the two drugs, and a weighted version of the midpoint would be part of the combination that causes the side effect. 

The key idea is a central-smoothing assumption: each hyperedge is called {\it central-smooth} if a weighted version of the midpoint of drug node representations is close enough to the representation of the side effect node. It is motivated by biological research that a side effect of a pair of drugs is caused by a combination of properties of the two drugs \cite{leucuta2006pharmacokinetics,corrie2011mechanisms}. Assuming that representations reflecting all properties of drugs are obtained, the midpoint (likewise, the sum) of two drug representations should contain all these properties of the two drugs. A weighted midpoint, which in the ideal case, would contain properties from each drug, represents a specific combination of the properties, potentially reflecting the cause of a side effect. The idea of summing representations to reflect a combination of features from two entities has been used in the past, such as in translation model for knowledge graph embedding (TransE, for directed graphs \cite{bordes2013translating}) or kernels for link prediction (pairwise kernels for undirected graphs \cite{basilico2004unifying}). %In the next parts, we present the central-smoothing assumption for each dimension, to construct the corresponding central-smoothing hypergraph Laplacian. 

\textbf{Central-smoothing measure on a hyperedge.} 
In the embedding space of $K$-dimension, considering dimension $k$  with  the embedding of nodes: $X_k \in \mathbb{R}^{|V|}$ that $X_{k,u} \in \mathbb{R}$ is the embedding of node $u\in V$. Given a hyperedge $e = (u,v,t)$, a weight $W_{k,t} \in \mathbb{R}^+$ is a parameter indicating the relevance of side effect $t$ on dimension $k$. We assign the  weight of side effect $t$ to the hyperedge ($w_k(e)= W_{k,t}$), and let $\mathbf{W}_k = diag(w_k(e_1), ..., w_k(e_{|E|}))$ be the diagonal matrix of the hyperedge weights. The central-smoothing measure on dimension $k$ of the hyperedge is defined as:
 
 \begin{align}\label{hsmootheq}
 ss^c(X_k, e) = W_{k,t}(\frac{X_{k,u} + X_{k,v}}{2} - X_{k,t})^2.
\end{align}

 \textbf{Central-smoothing measure on the hypergraph.} For hypergraph $G$, the central-smoothing measure on dimension $k$ is defined as the sum of the central-smoothing measures on all hyperedges:
 \begin{equation}\label{smootheq}
 sh^c(X_k,G) = \sum_{e \in E} W_{k,t} (\frac{X_{k,u} + X_{k,v}}{2} - X_{k,t})^2.  
 \end{equation}
%&= \sum_{e = (u, v, t) \in E^*} w_k(e) ss^c(x_k, e) \\

\textbf{Central-smoothing hypergraph Laplacian.}
Since $sh^c(X_k, G)$ is a nonnegative quadratic form, there exists a $\mathbf{L}_k\in \mathbb{R^{|V| \times |V|}}$ such that $  sh^c(X_k, G) = X_k^T \mathbf{L}_k X_k$. We call $\mathbf{L}_k$ as the \emph{central-smoothing hypergraph Laplacian}, which can be derived as follows.

Let $H \in \mathbb{R}^{|V| \times |E|}$ be a weighted oriented incidence matrix of $G$ that for a hyperedge $e \in E$, $H_{u,e} = H_{v,e} = \frac{1}{2}$ and $H_{t,e} = -1$, we have: 
\begin{align}\label{eq:clap0}
sh^c(X_k,G) &= \sum_{e \in E} W_{k,t} (\frac{X_{k,u} + X_{k,v}}{2} - X_{k,t})^2 \nonumber \\
&= X_k ^T H \mathbf{W}_k {H}^{\mathsf{T}} X_k \nonumber \\ 
&\stackrel{\text{def}}{=} X_k^T \mathbf{L}_k x_k. 
\end{align}

Then,
\begin{equation}\label{eq:clap}
    \mathbf{L}_k = H \mathbf{W}_k {H}^{\mathsf{T}}.
\end{equation}

% For $K$ dimensions, we have K central-smoothing Laplacian matrices: 
%\begin{align}
%\{L^c_k = H^c W_k{H^c}^T | k = 1...K\}.  \label{eq:lneww}
%\end{align}

\textbf{Computing the central-smoothing hypergraph Laplacian.}
The central-smoothing hypergraph Laplacian $\mathbf{L}_k$ in   \eqref{eq:clap} can be computed with the time complexity of $O(|E|)$. Concretely, each element $\mathbf{L}_{k,i,j}$ can be computed by:

\begin{align}
\mathbf{L}_{k,i,j} = \sum_{e \in E | i, j \in e} w_k(e)H_{i,e} H_{j, e}.
\end{align}
We have four cases:
\begin{itemize}
\item  $\mathbf{L}_{k,i,j} = \mathbf{L}_{k,j,i} = \frac{1}{4} \sum _{t \in V_s | (i, j, t) \in E} W_{k,t}$ if $i != j \in V_D$.
\item  $\mathbf{L}_{k,i,j} = \mathbf{L}_{k,j,i} = -\frac{1}{2} n_d(i,j)W_{k,j}$ if $i \in V_D, j \in V_S$.
\item  $\mathbf{L}_{k,i,i} = \frac{1}{4} \sum_{t | t \in V_S} m_d(i,t) W_{k,t}$ if $i  \in V_D$.
\item  $\mathbf{L}_{k,i,i} = q(i) W_{k,i}$ if $i \in V_S$.
\end{itemize}

 where $n_d(i,j) = | \{  (u, v, j) \in E | u= i \lor  v=i \}|$, $m_d(i,t) = | \{ u | (i, u, t) \lor (u,i,t) \in E\}|$, $q(i) = |\{ (u, v, i) | (u, v, i) \in E\} |$.

\textit{Complexity analysis.} Given $N$ convolution layers, the computational complexity for all central-smoothing hypergraph Laplacian is $O(N \cdot K \cdot |E|)$. Each $\mathbf{L}_k$ can be computed with a complexity of $O(|E|)$ by  iterating over all hyperedges in $E$ once, and for each hyperedge, the side effect weight is added to the corresponding elements in $\mathbf{L}_k$ and we have $N \times K$ Laplacian matrices to compute. 
We note that $K$ here is referred to the size of latent features, and this is not the original input features. In practice, even if the size of the original input features is very large, the number of latent features can be very small ($\le 200$), which is computationally tractable. 

\textbf{Non-weighted version.} In our experiments, we will examine the need for the weight of each side effect. So we here show a non-weighted version of central-smoothing hypergraph Laplacian, called $\mathrm{CentSimple}$ by fixing $\mathbf{W}_k$ to be an identity matrix, where
 the central-smoothing hypergraph Laplacian in \eqref{eq:clap} becomes:
\begin{align}
\tilde{\mathbf{L}}_k = H{H}^\mathsf{T}.
\end{align}

\subsection{Central-Smoothing Hypergraph Neural Networks (HGNNs)}

\textbf{Transforming input features to latent spaces}

We first transform the input feature vector of drugs and one-hot vector of side effects to the $K$-dimension latent space  by using a two-layer feedforward neural network for drugs, and a one-layer feedforward neural network (as an embedding table) for side effect, respectively, as follows: 
  \begin{align*}
X_D^{(0)} &= f_D(X_D)\\
X_S^{(0)} &= f_S(X_S),
\end{align*}
where $X_D \in \mathbb{R}^{|K_0| \times |V_D|}$ is the drug input features with feature size $K_0$, $X_S \in \mathbb{R}^{|V_S| \times |V_S|}$ is the one-hot vector of side effect, $X_D^{(0)} \in \mathbb{R}^{ K \times |V_D|}$,  $X_S^{(0)} \in \mathbb{R}^{ K \times |V_S|}$ and $f_D$ and $f_S$ are the corresponding feedforward neural networks.

\textbf{Convolution layers on the latent spaces}

 We adapt HGNN layers \cite{feng2019hypergraph} using $\mathbf{L}_k$ at dimension $k$. Given hypergraph Laplacian $\mathbf{L}_k$, we have the normalized adjacency matrix with a self-loop at each node:
 \begin{align}
   \tilde{A}_k = 2I - d_{\mathbf{L}_k}^{-1/2} \mathbf{L}_k d_{\mathbf{L}_k}^{-1/2},   
 \end{align}
 where $d_{\mathbf{L}_k}$ is the degree matrix, corresponding to Laplacian $\mathbf{L}_k$ and $I$ is the identity matrix.
 
 Let $\tilde{D}_k$ be the corresponding degree matrix of $\tilde{A}_k$, each layer of  central-smoothing HGNNs has the following form:
 
 \begin{align}\label{onelayer}
     X^{(l+1)} = \sigma(\tilde{X}^{(l+1)} \Theta^{(l)}),
 \end{align}
 where $\tilde{X}^{(l+1)} = [\tilde{x}_1^{(l+1)}, ..., \tilde{x}_K^{(l+1)}]$ and $\tilde{x}_k^{(l+1)} = \tilde{D}_k^{-1/2} \tilde{A}_k \tilde{D}_k^{-1/2}x_k^{(l)}$, $\Theta^{(l)} \in \mathbb{R}^{K\times K}$ is the parameters for the transformation from layer $(l)$ to layer $(l+1)$, and $\sigma$ is an activation function.

\subsection{Predicting Drug-Drug Interactions}
Assuming that ${X^*}^{\mathsf{T}} \in \mathbb{R}^{|V| \times K}$ is the final node representation 
%in hypergraph $G$ 
with learnt weights $W^* = \{W_k^* | k = 1...K\}$. For all $e=(u,v,t)$, $t$ is predicted to be a side effect of $u$ and $v$ if the representation of $t$ is close enough to the weighted midpoint of the two drug node representations (computed by score function $p(e, X^*, W^*)$). First, we compute smoothness measures $ssa(e, X^*, W^*)$ of $(u,v,t)$ on all dimensions:
\begin{align}
 ssa(e,X^*, W^*) = \sum_{k = 1}^K W^*_{k,t}(\frac{X^*_{k,u} + X^*_{k,v}}{2} - X^*_{k,t})^2.
\end{align}
Then, the prediction score is defined to be:
\begin{align}
p(e, X^*, W^*) = \frac{1}{1 + ssa(e,X^*, W^*)}.
\end{align}
If $p(e, X^*, W^*) > h$, a predefined threshold, then $t$ is predicted to be a side effect of $u$ and $v$.

\subsection{Objective Function of CentSmoothie}
Let $\bar{E} = V_D \times V_D \times V_S \setminus E$ be complement of the hyperedge set. The objective function to train $\mathrm{CentSmoothie}$ is to maximize the score $p(e,X^*, W^*)$ of the known hyperedges and minimize the score of the complement set $\bar{E}^*$. Then the objective function can be defined as:

\begin{align}\label{objective}
\min_{ W^* \ge 0, X^*} f(X^*, W^*) & = \sum_{e \in E} (1 - p(e, X^*, W^*))^2  \\ &+\lambda \sum_{e \in \bar{E}} p(e, X^*, W^*)^2, 
\end{align}
where $\lambda$ is a hyperparameter.

In practice, as $|\bar{E}|$ is too large, we randomly sample a subset of $\Omega \subset \bar{E}, |\Omega| = |E|$ to replace $\bar{E}$ in the objective function to reduce the computational cost (A CentSmoothie implementation available at \url{https://github.com/anhnda/CentSmoothieCode}). To keep the non-negative constraint on $W^*$, we used a projected gradient descent \cite{lin2007projected}.

\section{Experiments}

We conducted experiments to evaluate the performance of our proposed method, $\mathrm{CentSmoothie}$, a hypergraph neural network with a central-smoothing assumption, in two scenarios: (i) a synthetic dataset and (ii)  three real DDI datasets. On the synthetic dataset, we aimed to validate that  $\mathrm{CentSmoothie}$ could achieve higher performances than traditional hypergraph neural networks, by using the data generated from the central-smoothing assumption. On the real DDI datasets, we  examined the performance of $\mathrm{CentSmoothie}$ in comparison with baseline models, to prove that the central-smoothing assumption is suitable for DDI data. 

For both scenarios, we used 20-fold cross-validation using the mean  AUC (area under the ROC curve) and the mean AUPR (area under the precision-recall curve) with standard deviations, to validate the prediction performances \cite{zitnik2018modeling}. 

%We conducted experiments to evaluate the performance of our proposed method, $\mathrm{CentSmoothie}$, a hypergraph neural network with a central-smoothing assumption on three real DDI datasets. We examine the performance of $\mathrm{CentSmoothie}$ in comparison with baseline models, to prove that the central-smoothing assumption is suitable for DDI data \footnote{Experiments on synthetic data can be found at the supplement.}. 

%We used 20-fold cross-validation using the mean  AUC (area under the ROC curve) and the mean AUPR (area under the precision-recall curve) with standard deviations, to validate the prediction performances \cite{zitnik2018modeling}. 

For graph and hypergraph neural networks, the numbers of layers and the embedding sizes were in [1, 2, 3] and [10, 20, 30], respectively. The activation function was rectified linear unit (ReLu). The hyperparameter $\lambda$ was fixed: 0.01. The results obtained were the highest performances with the number of layers of 2 and the embedding size of 20 for all methods. All experiments were run on a computer with Intel Core I7-9700 CPU, 8 GB GeForce RTX 2080 GPU, and 32 GB RAM.

\subsection{Synthetic Data}

\subsubsection{Generation}
We generated a synthetic dataset with the idea that each drug has several groups of features and the combination of two groups of features  leads to a side effect of the drugs. We fixed the number of drugs $D=500$, the number of side effects: $S=45$, and changed maximum number of groups of drug features from $1$ to $6$. The detail of the generation process can be found in the supplement.

\subsubsection{Comparing Methods}

For the synthetic dataset, we used the  central-smoothing hypergraph neural networks  $\mathrm{CentSmoothie}$, the non-weighted central-smoothing hypergraph neural networks $\mathrm{CentSimple}$, and  the existing spectral based hypergraph neural network, $\mathrm{HPNN}$ \cite{feng2019hypergraph}.

\subsubsection{Results}

\begin{figure}[tp]
%\centering
\subfloat[][]{\includegraphics[width=0.5\linewidth]{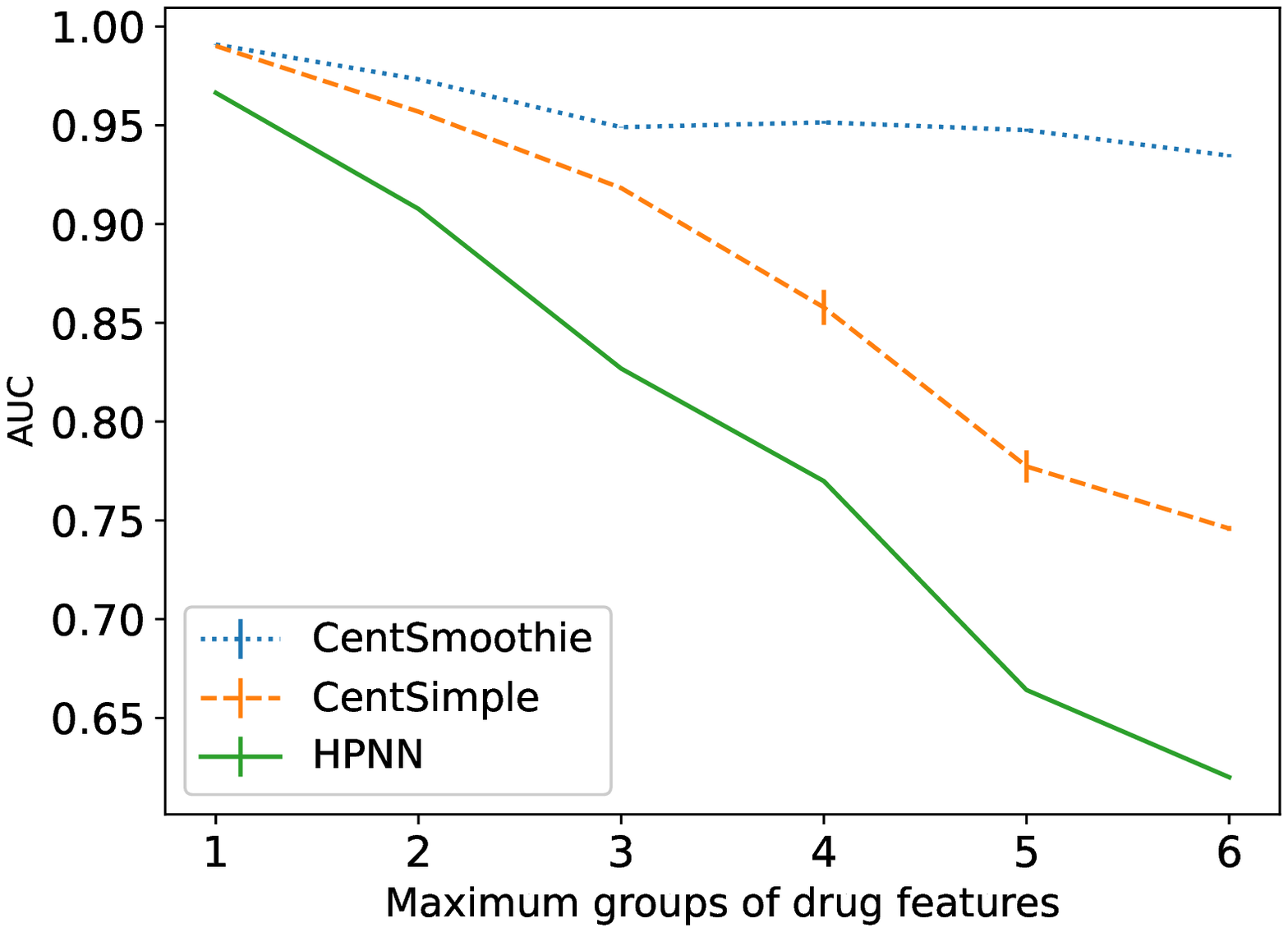}\label{fig:synauc}}
\subfloat[][]{\includegraphics[width=0.5\linewidth]{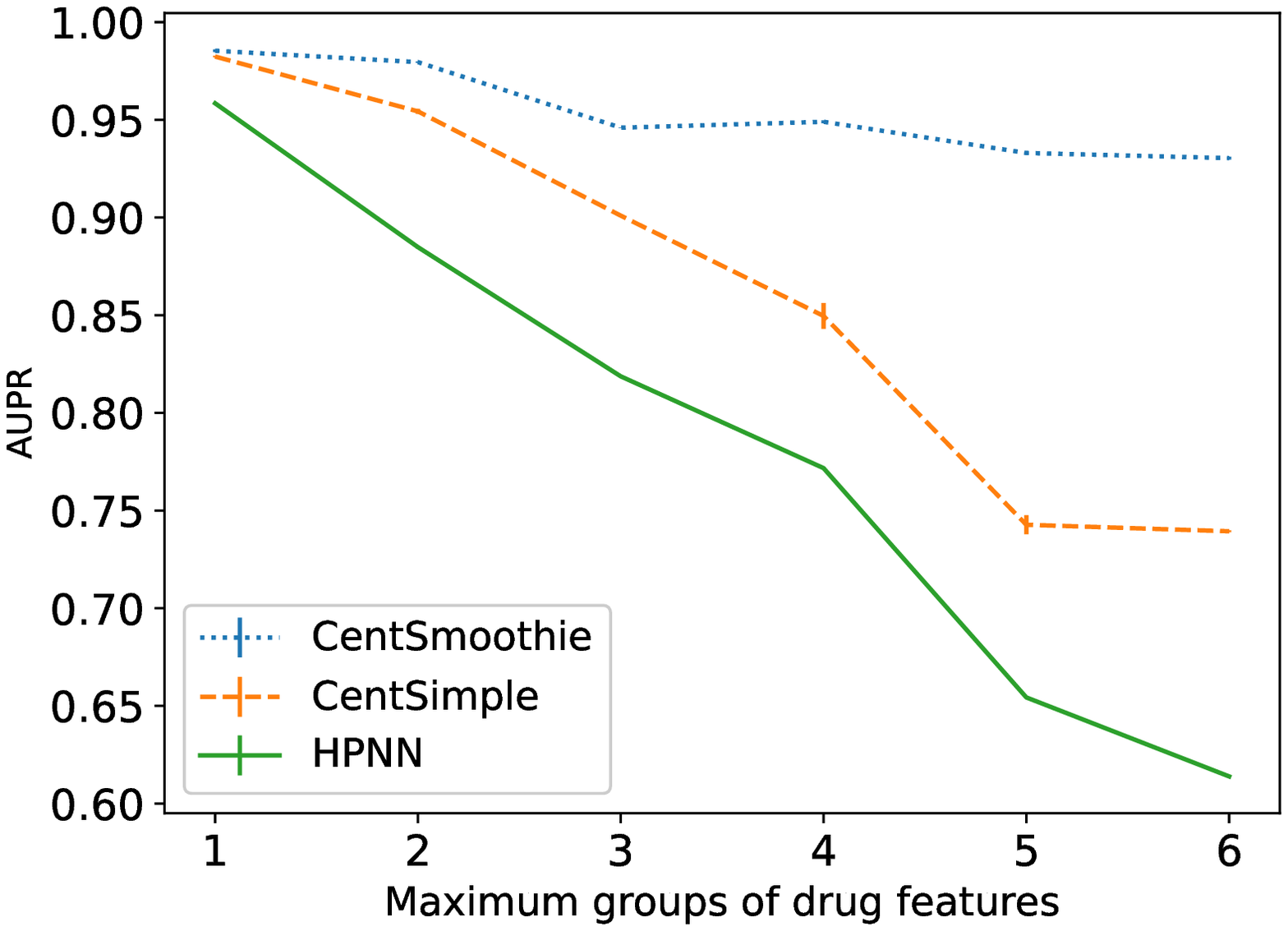}\label{fig:synaupr}}

% \subfloat[][]{\includegraphics[width=\linewidth]{figs/SYN_AUC.eps}\label{fig:synauc}}
  \caption{Synthetic data performance comparison: (a) AUC and (b) AUPR.}
  %s of HPNN, $\mathrm{CentSimple}$, and $\mathrm{CentSmoothie}$ on the synthetic data using (a) AUC and (b) AUPR.}
\label{fig:syn}
\end{figure}

Fig. \ref{fig:syn} shows the AUC and AUPR of each compared method, obtained by changing maximum number of groups of features for drugs.
We could easily see that $\mathrm{CentSmoothie}$ achieved the highest AUC and AUPR scores for all values of x-axis, followed by $\mathrm{CentSimple}$ and then $\mathrm{HPNN}$.
In particular, the AUC scores of $\mathrm{CentSmoothie}$ were always  higher than 0.95, while those of HPNN decreased when drugs are more complex with larger numbers of groups drugs features. This clearly showed that $\mathrm{CentSmoothie}$ could correctly capture the patterns generated by the central smoothing assumption, particularly for larger numbers of groups of drug features. Similarly, the AUC scores of $\mathrm{CentSimple}$ decreased with higher number of maximum number of groups of features, e.g. around 0.75 at 6. The pattern for AUPR scores was also similar to that of AUC scores.
 This result showed that $\mathrm{CentSmoothie}$ could learn different side effects for drug pairs more effectively than $\mathrm{CentSimple}$, implying the significance of using a weight for each side effect in $\mathrm{CentSmoothie}$.  

\subsection{Real Data}

\subsubsection{Data description}

We used three real DDI datasets: TWOSIDES, CADDDI, and JADERDDI. TWOSIDES is a public dataset for DDI extracted from the FDA adverse event reporting system (US database) \cite{tatonetti2012data}. To our knowledge, TWOSIDES is the largest and commonly used benchmark dataset for DDI \cite{zitnik2018modeling,rohani2019drug,xu2019mr}. 
 In a similar manner as in \cite{tatonetti2012data} of TWOSIDES, we used significant tests to generate two new DDI datasets: CADDDI from Canada vigilance adverse reaction report (Canada database, from 1965 to Feb 2021) \cite{cader}  and JADERDDI from The Japanese Adverse Drug Event Report (Japanese database, from 2004 to March 2021) \cite{jader}.
We only selected small molecular drugs appearing in DrugBank \cite{wishart2018drugbank}. Each drug feature vector was a binary vector with the size of 2,329, indicating the existences of 881 substructures and 1,448 interacting proteins \cite{nguyen2021survey}.
The statistics of the final datasets is shown in Table \ref{tb:realData}.

\begin{table*}[htbp]
\centering
\caption{Statistics of the three real datasets.}
\label{tb:realData}
%\begin{tabular}{|c|c|c|c|c|c|c|c|c|c|c|c|c|c|c|c|}
\begin{tabular}{lrrrrrrrrrrrrrrrrrrrr}
\hline
\multirow{2}{*}{{Dataset}} &\multirow{2}{*}{{\#drugs}} & \multirow{2}{*}{{\#side effects}} & \multirow{2}{*}{{\#drug-drugs}} & \multirow{2}{*}{{\#drug-drug-side effects}} & \multirow{2}*{{Avg. side effects/drug-drugs}} & \multicolumn{3}{c}{{drug-drugs/ side effects}} \\ \cline{7-9}
& & & & & & {\textit{Min}} & {\textit{Max}} & {\textit{Avg}}\\ \hline
TWOSIDES & 557 & 964 & 49,677 &  3,606,046 & 72.58  & 288& 22,520 & 3740.7 \\ %\hline
CADDDI & 587 & 969 & 21,918 & 373,976 & 17.06 & 89 & 3288 & 385.9 \\ %\hline
JADERDDI & 545 & 922 & 36,929 & 222,081 & 6.01 & 60 & 1922 & 240.9 \\ \hline
\end{tabular}
\end{table*}

\begin{table*}[htbp]
\centering
\caption{Comparison of performances of the methods on the real DDI datasets.}
\label{tb:onreal}
%\begin{tabular}{|c|c|c|c|c|c|c|} 
\begin{tabular}{lrrrrrrrrrrrrrr} 
\hline
\multirow{2}{*}{{Method}} & \multicolumn{2}{c}{{{TWOSIDES}}} &  \multicolumn{2}{c}{{{CADDDI}}} &  \multicolumn{2}{c}{{{JADERDDI}}} \\ \cline{2-7}
 & {\textit{AUC}} & {\textit{AUPR}} & {\textit{AUC}} & {\textit{AUPR}} & {\textit{AUC}} & {\textit{AUPR}} \\ \hline
 MLNN & 0.8372 $\pm$ 0.0050 & 0.7919 $\pm$ 0.0041 & 0.8689 $\pm$ 0.0021 & 0.6927 $\pm$ 0.0082 & 0.8578 $\pm$ 0.0015 &  0.3789 $\pm$ 0.0020 \\ %\hline
 MRGNN &  0.8452 $\pm$ 0.0036 & 0.8029 $\pm$ 0.0039  & 0.9226 $\pm$ 0.0015 & 0.7113 $\pm$ 0.0031 & 0.9049 $\pm$ 0.0009 & 0.3698 $\pm$ 0.0019 \\ %\hline
 Decagon & 0.8639 $\pm$ 0.0029  & 0.8094 $\pm$ 0.0024 & 0.9132 $\pm$ 0.0014 & 0.6338 $\pm$ 0.0029 &  0.9099 $\pm$  0.0012 &  0.4710 $\pm$ 0.0027  \\ %\hline
SpecConv & 0.8785 $\pm$ 0.0025 &  0.8256$\pm$  0.0022 & 0.8971 $\pm$ 0.0055 & 0.6640 $\pm$ 0.0014 & 0.8862 $\pm$ 0.0025 & 0.5162 $\pm$ 0.0047 \\ %\hline
HETGNN & 0.9113 $\pm$ 0.0004 & 0.8267 $\pm$ 0.0005 & 0.9371 $\pm$  0.0004 & 0.7974 $\pm$  0.0011 & 0.8989 $\pm$ 0.0007 & 0.5618 $\pm $ 0.0012 \\
HPNN & 0.9044 $\pm$0.0003 & 0.8410 $\pm$  0.0007 & 0.9495 $\pm$ 0.0004 & 0.7020 $\pm$ 0.0018 & 0.9127 $\pm$ 0.0004 & 0.5198 $\pm$ 0.0016 \\ %\hline
$\mathrm{CentSimple}$ &  0.9242 $\pm$  0.0003  & 0.8638 $\pm$  0.0011 & 0.9584 $\pm$ 0.0005 & 0.6890 $\pm$ 0.0016 & 0.9239 $\pm$ 0.0007 & 0.5349 $\pm$ 0.0021 \\ %\hline
$\mathrm{CentSmoothie}$ & \textbf{0.9348} $\pm$ 0.0002 &  \textbf{0.8749} $\pm$  0.0013 &  \textbf{0.9846} $\pm$ 0.0001 & \textbf{0.8230} $\pm$ 0.0019 & \textbf{0.9684} $\pm$ 0.0004 & \textbf{0.6044} $\pm$ 0.0025 \\ \hline
\end{tabular}
\end{table*}

\subsubsection{Comparing Methods}

On the real datasets, we compared our proposed methods to baselines:  none-graph based, graph based, and hypergraph based methods. For the none-graph based method, we used a multi-layer feedforward neural network (MLNN) \cite{rohani2019drug}.  For graph neural networks, on the drug molecular graphs, we used $\mathrm{MRGNN}$ \cite{xu2019mr} with the recommended hyperparameter settings. On the DDI graph, we used Decagon \cite{zitnik2018modeling}, a spatial convolution, SpecConv (a spectral convolution graph neural networks)  \cite{kipf2016semi}, and HETGNN (a heterogeneous graph neural network) \cite{zhang2019heterogeneous}. 
For hypergraph neural networks, we used the existing spectral convolution hypergraph neural network, $\mathrm{HPNN}$ \cite{feng2019hypergraph}. We also showed the results of  $\mathrm{CentSimple}$ to see the effect of central-smoothing without having weights for side effects.

\begin{figure}[tp]
%\centering
\subfloat[][]{\includegraphics[width=0.9\linewidth]{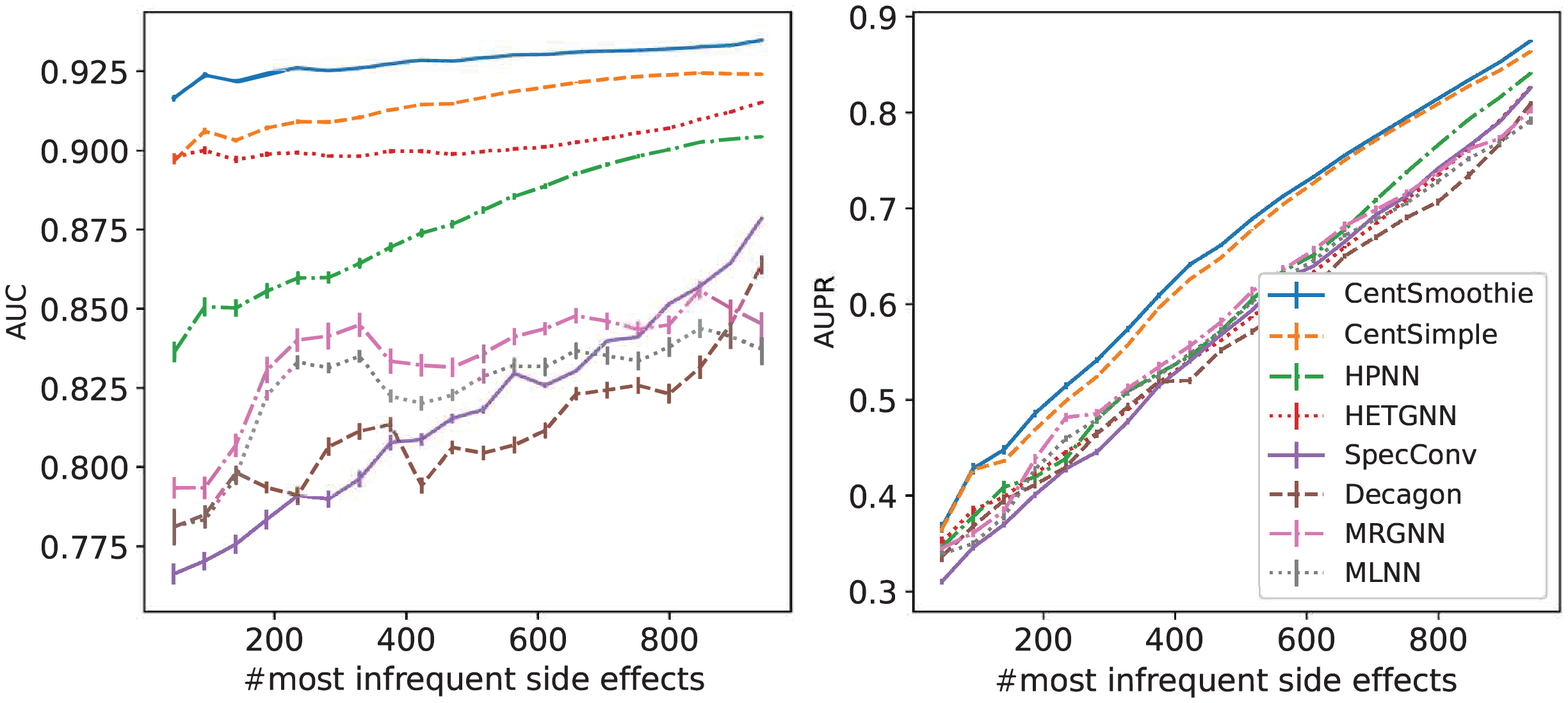}\label{fig:ar}}

\vspace{-2mm}
  \subfloat[][]{\includegraphics[width=0.9\linewidth]{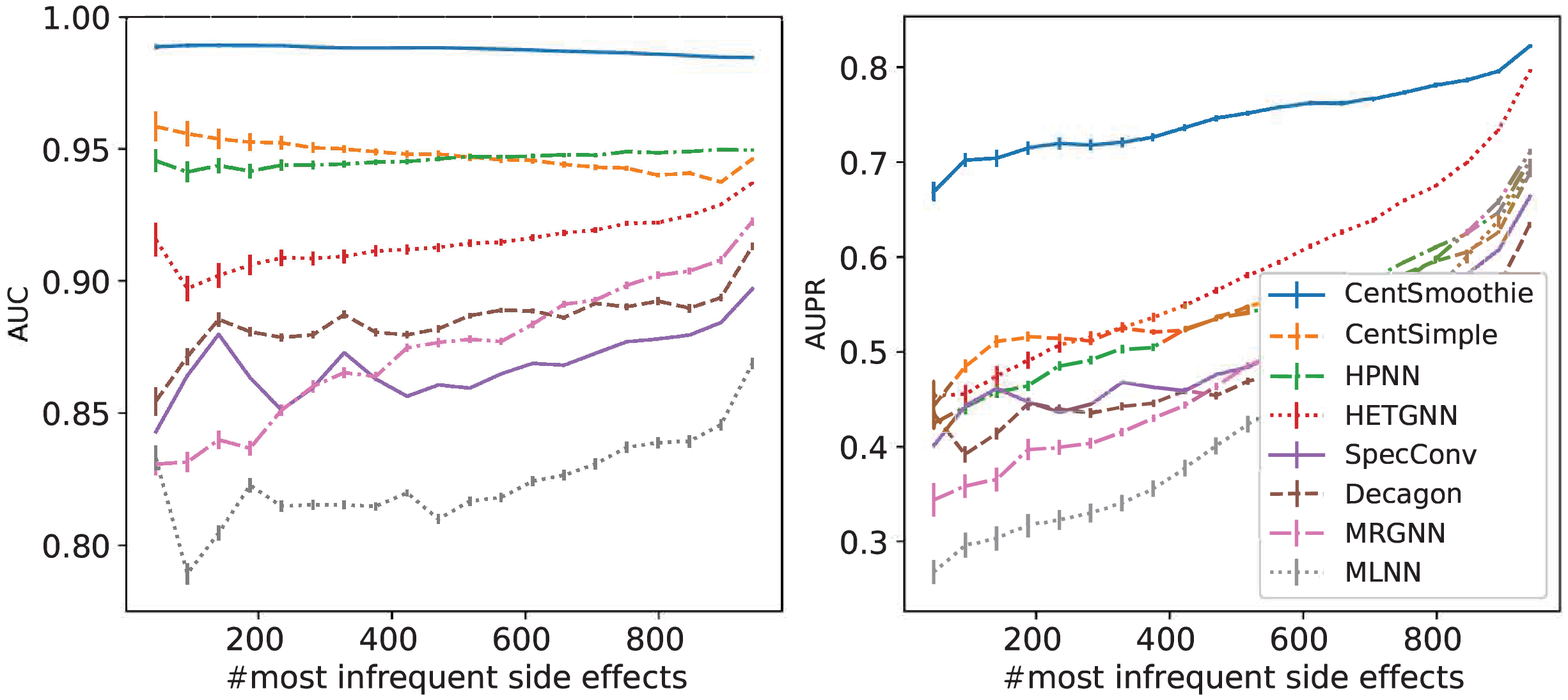}\label{fig:cr}}

\vspace{-2mm}
  \subfloat[][]{\includegraphics[width=0.9\linewidth]{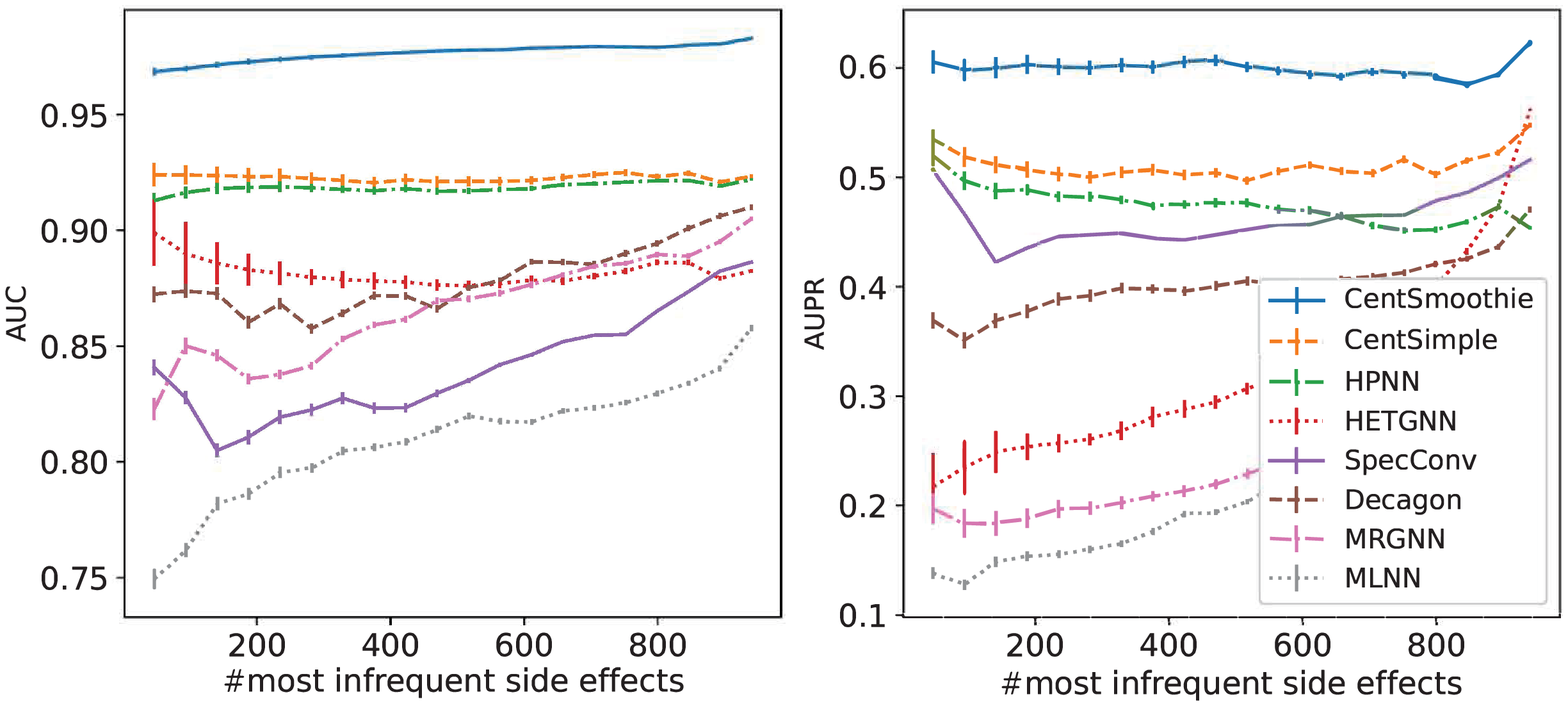}\label{fig:jr}}
  \caption{Performance comparison (AUC (left) and AUPR (right)) on (a) TWOSIDES, (b) CADDDI and (c) JADDERDDI.}
  \label{fig:real}
\end{figure}

%  with none-graph base methods (MLNN), graph neural networks on drug molecular graphs (MRGNN), graph neural networks on DDI graph (SpecConv, Decagon, HETGNN), existing hypergraph neural network (HPNN), $\mathrm{CentSimple}$ and $\mathrm{CentSmoothie}$  using AUC (left) and AUPR (right). }

\subsubsection{Results}

Table \ref{tb:onreal} shows the AUC  scores and AUPR scores of all methods. We could see that again $\mathrm{CentSmoothie}$ achieved the highest AUC and AUPR scores in all three datasets. For TWOSIDES, $\mathrm{CentSmoothie}$ achieved 0.9348 in AUC and 0.8749 in AUPR, followed by  $\mathrm{CentSimple}$ (0.9242 and 0.8638), HPNN (0.9044 and 0.8410), HETGNN (0.9113 and 0.8267),  SpecConv (0.8785 and 0.8256), Decagon (0.8639 and 0.8094), MRGNN (0.8452 and 0.8029), and MLNN (0.8372 and 0.7919). 

For CADDDI and JADERDDI, $\mathrm{CentSmoothie}$ had the highest performances with AUC and AUPR: (0.9845 and 0.8230) and (0.9684 and 0.6044), respectively. The second and third best methods were $\mathrm{CentSimple}$ and HPNN, respectively. 

In particular, in AUC, there existed two clear performance gaps. The first one was between hypergraph based methods ($\mathrm{CentSmoothie}$, $\mathrm{CentSimple}$ and HPNN) and non-hypergraph based methods (HETGNN, SpecConv, Decagon, MRGNN, and MLNN). The second one was between $\mathrm{CentSmoothie}$ and ($\mathrm{CentSimple}$ and HPNN).
The first gap showed the advantage of using hypergraph based method for predicting drug-drug interaction. The second gap showed the advantage of central smoothing over regular smoothing. In addition, we could see the importance of learning weights for each side effect to improve the prediction performance.

In AUPR, there was a clear gap between $\mathrm{CentSmoothie}$ and the remaining methods. This again showed the advantage of learning weights under the central smoothing assumption for predicting DDI.

$\mathrm{CentSmoothie}$ can learn the representations of side effects together with drugs to leverage the relationships of side effects (see the supplement for representation visualization of side effects).
%for some infrequent side effects learnt by CentSmoothie can be found in the supplement.}.
These side effect representations might be useful for infrequent side effects which are harder to predict due to the scarcity of positive training data. Fig. \ref{fig:real} showed the AUC (left) and AUPR (right) scores of the methods on the subset of most infrequent side effects, obtained by starting with the most infrequent side effect and adding the next infrequent side effects to the subset. From both AUC and AUPR scores in Fig. \ref{fig:real}, we could see that $\mathrm{CentSmoothie}$ achieved the best performances  for all values of x-axis (the rightmost point of x-axis corresponds to using all side effects), being followed by $\mathrm{CentSimple}$ and HPNN.

\section{Conclusion}

We have presented $\mathrm{CentSmoothie}$, a hypergraph neural network, for predicting drug-drug interactions, to learn representations of side effects together with drug representations in the same space. 
A unique feature of $\mathrm{CentSmoothie}$ is a new central-smoothing formulation, which can be incorporated into the hypergraph Laplacian, to model drug-drug interactions. % of the hypergraph neural network.
Our extensive experiments of using both synthetic and three real datasets confirmed clear performance advantages of $\mathrm{CentSmoothie}$ over existing hypergraph and graph neural network methods, indicating that $\mathrm{CentSmoothie}$ could learn representations of drugs and side effects simultaneously with the central-smoothing assumption. 
Furthermore, $\mathrm{CentSmoothie}$ kept high performance on the  infrequent side effects for which the performances of other methods dropped significantly, indicating that $\mathrm{CentSmoothie}$ allows leveraging the relationships among side effects to help the difficult cases of less frequent side effects.
For future work, it is interesting to extend the central-smoothing assumption into more general cases not limiting to 3-uniform hypergraphs. In addition, learning adaptive ratios to replace the constraint of the midpoint might be considered.

\bibliographystyle{IEEEtran}
\bibliography{ref.bib}

\renewcommand\appendixname{Supplements}

\cleardoublepage
\newpage

\appendix

\section*{Derivation of incidence matrix H}
We show that given $H \in \mathbb{R}^{|V| \times |E|}$, where $H_{u,e} = H_{v,e} = \frac{1}{2}$ and $H_{t,e} = -1$ for each $e=(u,v,t) \in E$ then:

\begin{align*}\label{eq:clap0}
\sum_{e \in E} W_{k,t} (\frac{X_{k,u} + X_{k,v}}{2} - X_{k,t})^2 \nonumber = X_k ^T H \mathbf{W}_k {H}^{\mathsf{T}} X_k \nonumber 
\end{align*}
where $X_k \in \mathbb{R}^{|V| \times 1}$, $X_{k,u}$ be the corresponding value of $u$ in $X_k$, $\mathbf{W}_k = diag(w_{k}(e_1), ..., w_{k}( e_{|E|}))$ with $w_{k}(e) = W_{k,t}$.

Proof: 
Let $H_{.,e} \in \mathbb{R}^{|V| \times 1}$ be the column of $H$ corresponding to hyperedge $e$. We have:
\begin{align*}
    & \sum_{e \in E} W_{k,t} (\frac{X_{k,u} + X_{k,v}}{2} - X_{k,t})^2 \\
     &=  \sum_{e \in E} (\frac{X_{k,u} + X_{k,v}}{2} - X_{k,t}) W_{k,t}  (\frac{X_{k,u} + X_{k,v}}{2} - X_{k,t}) \\
     &=  \sum_{e \in E} (X_k H_{.,e}) W_{k,t}  (X_k H_{.,e}) \\
     &= X_k^\mathsf{T} H \mathbf{W}_k H^T X_k  \hspace{20pt} \square
\end{align*}

\section*{Computing $L_k$}
Given the formulation for $L_k$:
\begin{align*}
\mathbf{L}_{k,i,j} = \sum_{e \in E | i, j \in e} w_k(e)H_{i,e} H_{j, e}.
\end{align*}
We have four cases:

\begin{enumerate}
    \item  $i, j \in V_D$, $i!=j$, meaning that $H_{i,e} = H_{j, e} = \frac{1}{2}$, hence:
    \begin{align*}
        \mathbf{L}_{k,i,j} &= \sum_{e \in E | i, j \in e} w_k(e)H_{i,e} H_{j, e} \\
                           &= \frac{1}{4}  \sum_{e \in E | i, j \in e}  w_k(e) \\
                           &= \frac{1}{4}  \sum_{t \in V_S | e=(i,j,t) \in E} W_{k,t}
    \end{align*}
    
     \item $i \in V_D, j \in V_S$, meaning that $H_{i,e} = \frac{1}{2}$ and $H_{j,e} = -1$, hence:
     
     \begin{align*}
         \mathbf{L}_{k,i,j} &= \mathbf{L}_{k,j,i} = \sum_{e \in E | i, j \in e} w_k(e)H_{i,e} H_{j, e} \\
         &= \frac{-1}{2} \sum_{e \in E | i, j \in e} w_k(e) = \frac{-1}{2} \sum_{e \in E | i, j \in e} W_{k,j}  \\
         &=  \frac{-1}{2}  W_{k,j} \sum_{e \in E | i, j \in e} 1 \\
         &= \frac{-1}{2}  W_{k,j} \sum_{e=(u,v,j) \in E | u= i \lor  v=i } 1 \\
         &= \frac{-1}{2}  W_{k,j} n_d(i,j)
     \end{align*}
     where  $n_d(i,j) = | \{  (u, v, j) \in E | u= i \lor  v=i \}|$.
     
     \item $i = j \in V_D$, $H_{i,e} = H_{j, e} = \frac{1}{2}$, hence:
     \begin{align*}
          \mathbf{L}_{k,i,i} & = \sum_{e=(u,v,t) \in E|  u= i \lor  v=i } w_k(e)H_{i,e} H_{i, e} \\
          &= \frac{1}{4}  \sum_{e=(u,v,t) \in E|  u= i \lor  v=i } w_{k,t}\\
          &= \frac{1}{4}  \sum_{t \in V_S} w_{k,t} \sum_{e=(u,v,t) \in E|  u= i \lor  v=i } 1 \\
          &=  \frac{1}{4}  \sum_{t \in V_S} w_{k,t} m_d(i,t)
     \end{align*}
     where $m_d(i,t) = | \{ u | (i, u, t) \lor (u,i,t) \in E\}|$.
     \item $i=j \in V_S$, meaning that $H_{i,e} = H_{j, e} = -1$, hence:
     \begin{align*}
          \mathbf{L}_{k,i,i} & = \sum_{e=(u,v,t) \in E|  u= i \lor  v=i } w_k(e)H_{i,e} H_{i, e} \\
          &= \sum_{e=(u,v,i) \in E } w_k(e) = W_{k,i} \sum_{e=(u,v,i) \in E } 1 \\
          &=   W_{k,i} q(i)
     \end{align*}
     where  $q(i) = |\{ (u, v, i) | (u, v, i) \in E\} |$.
\end{enumerate}

\section*{Detail of synthetic data generation}
%\subsubsection{Synthetic data generation}

%We generated a synthetic dataset with the idea that each drug has several groups of features and the combination of two groups of features  leads to a side effect of the drugs. We fixed the number of drugs $D=500$, the number of side effects: $S=45$, and changed maximum number of groups of drug features from $1$ to $6$.
The idea to generate synthetic data is that each drug has several groups of features and the combination of two groups of features  leads to a side effect of the drugs. The generation process consists of three steps: 
\begin{itemize}
\item Step 1: Generating groups of features and their combinations. Suppose that there were $n$  groups of features: $G = \{ g_1,..., g_n\}$. There are maximally $\frac{n(n-1)}{2}$ group combinations: $P = \{ (g_i, g_j) | i = 1...n, j= i+1 ...n\}$. Each group combination $p_i \in P, i = 1...|P|$ is assigned with a side effects $s_i$. 

\item Step 2: Generating drug features. Let $a$ be the number of features in a group, $D$ be the number of drugs, and  $m$ be the maximum number of groups of features for each drug. 

For each drug $i$, we first uniformly sampled the number of groups $1 \le n_i \le m$ and then sample $n_i$ groups from $G$. Let $G_i \in G$ be the sampled groups of drug $i$. Let the binary vector $\textbf{b}_{i} \in \mathbb{R}^{a.n}$ indicated the existence of features for drug $d_i$ that $\mathbf{b}_{i}(j) = 1$ if $ \left\lfloor  j /  a \right\rfloor \in G_i$, otherwise $ \mathbf{b}_{i}(j)  = 0$.

The feature vector of drug $i$ was sampled from a Gaussian distribution with mean $\mathbf{b}_{i}$ and variance $\sigma$: $\mathbf{f}_{i} = Gaussian( \mathbf{b}_{i}, \sigma)$.

\item Step 3: Generating triples of drug-drug and side effects. For each pair of two drugs generated from Step 2, we matched the group combinations of the two drugs with the corresponding side effects from Step 1. 
For a pair of two drugs $i$ and $j$ with corresponding groups $G_i$ and $G_j$, let $P_{ij} = G_i \times G_j$ and $S_{ij} = \{ s_t  | p_t \in P_{ij} \}$, we generated the triples: $ E_{ij} = \{ (d_i, d_j, s_t) | s_t \in S_{ij} \}$.

\end{itemize}

By going through all pairs of drugs, we obtained the synthetic data set with the drug feature vectors $F = \{ f_i | i = 1...n\}$ and the triples of drug-drug-side effect $ E = \cup_{i = 1...n, j = i+1...n} E_{ij}$.

 We set the number of groups $n=10$, the number of features in each group $a=3$, the variance $\sigma=0.01$, and the number of drugs $D=500$. We changed $m$ in the range of $[1, 2, \cdots 6]$.

\section*{Details of experiments}
\subsection*{Extracting new datasets}
For Canada vigilance adverse and JADERDDI from The Japanese Adverse Drug Event Report, each database consists of reports such that each report contains drugs and the corresponding observed side effects of a patient. 

The extraction from these databases was that for each drug pair, we divided the reports into two groups: an exposed group for the reports having the drug pair and a nonexposed group for the reports not having the drug pair.  Then, for each side effect, Fisher's exact test with the threshold p-value of 0.05 was used to check if the occurrence rate of the side effect in the exposed group was significantly higher than in the nonexposed group.

Finally, we obtained a set of significant triples of drug-drug-side effects for each database.

Regarding the overlapping of the datasets, between TWOSIDES and CADDDI, there is 24.8\% of overlapping in side effect names and 59.8\% of overlapping in drug names. For JADERDDI, we used Google service to translate Japanese drug names to English, mostly written in Katakana, which are more reliable to translate. The overlapping in drug names of TWOSIDE and JADERDDI is 15\%. We did not calculate the overlapping of side effects in JADDERDDI since the side effect names were not translate

\subsection*{Splitting data and hyperparameter selection}
We split the significant triples (positive set) of drug-drug-side effects into 20 folds with the same ratios for side effects in all folds. The negative set is the complement set of the positive set, defined by: $V_D \times V_D \times V_S / E$. 

We ran the methods with all hyperparameters in the grid search range. For each method, we selected the hyperparameters having the highest mean value of the 20-fold cross-validation.

\begin{figure*}[tp]
\centering
\subfloat[][]{\includegraphics[width=0.3\textwidth]{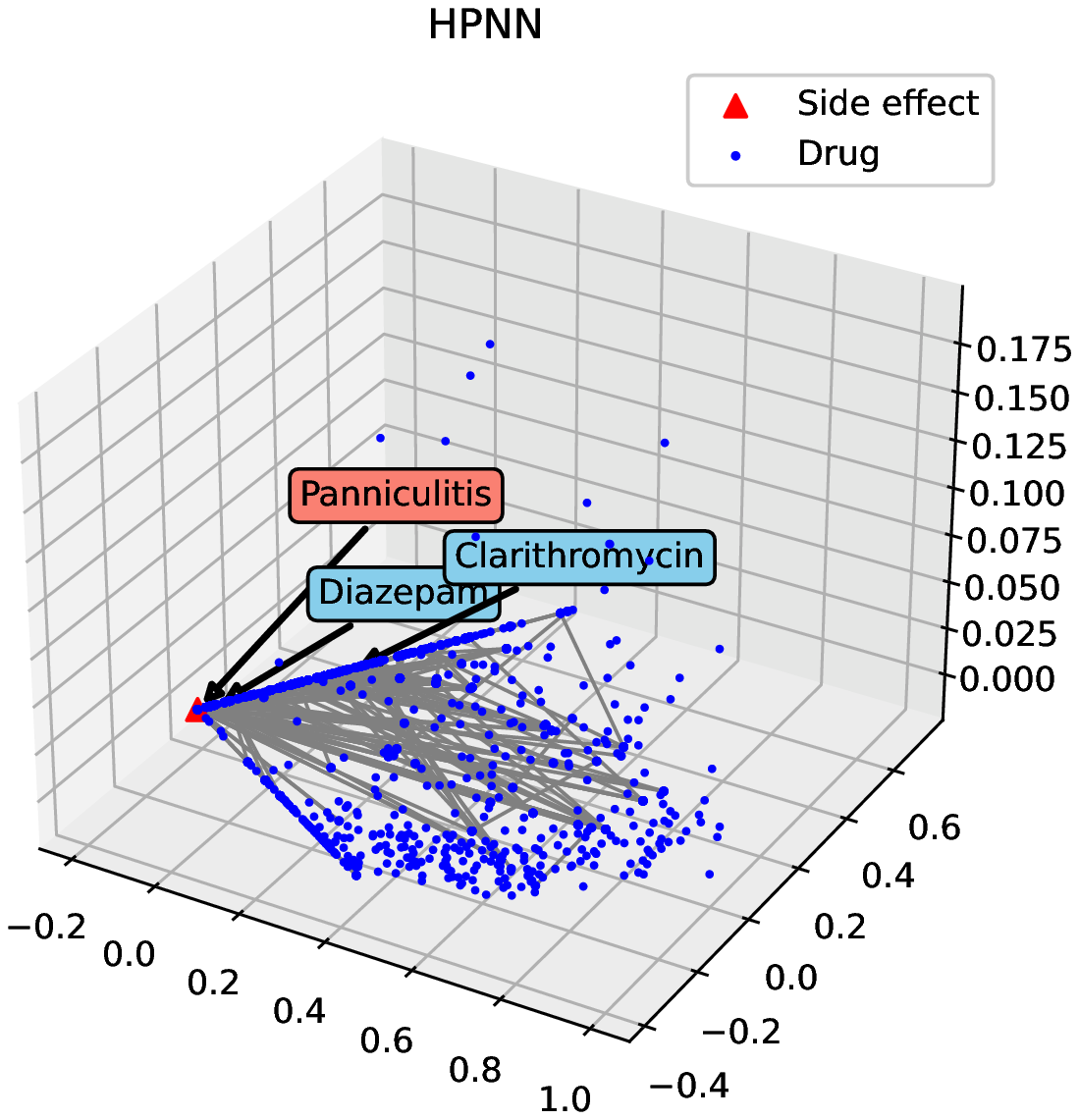}\label{fig:oldV1}}
\subfloat[][]{\includegraphics[width=0.34\textwidth]{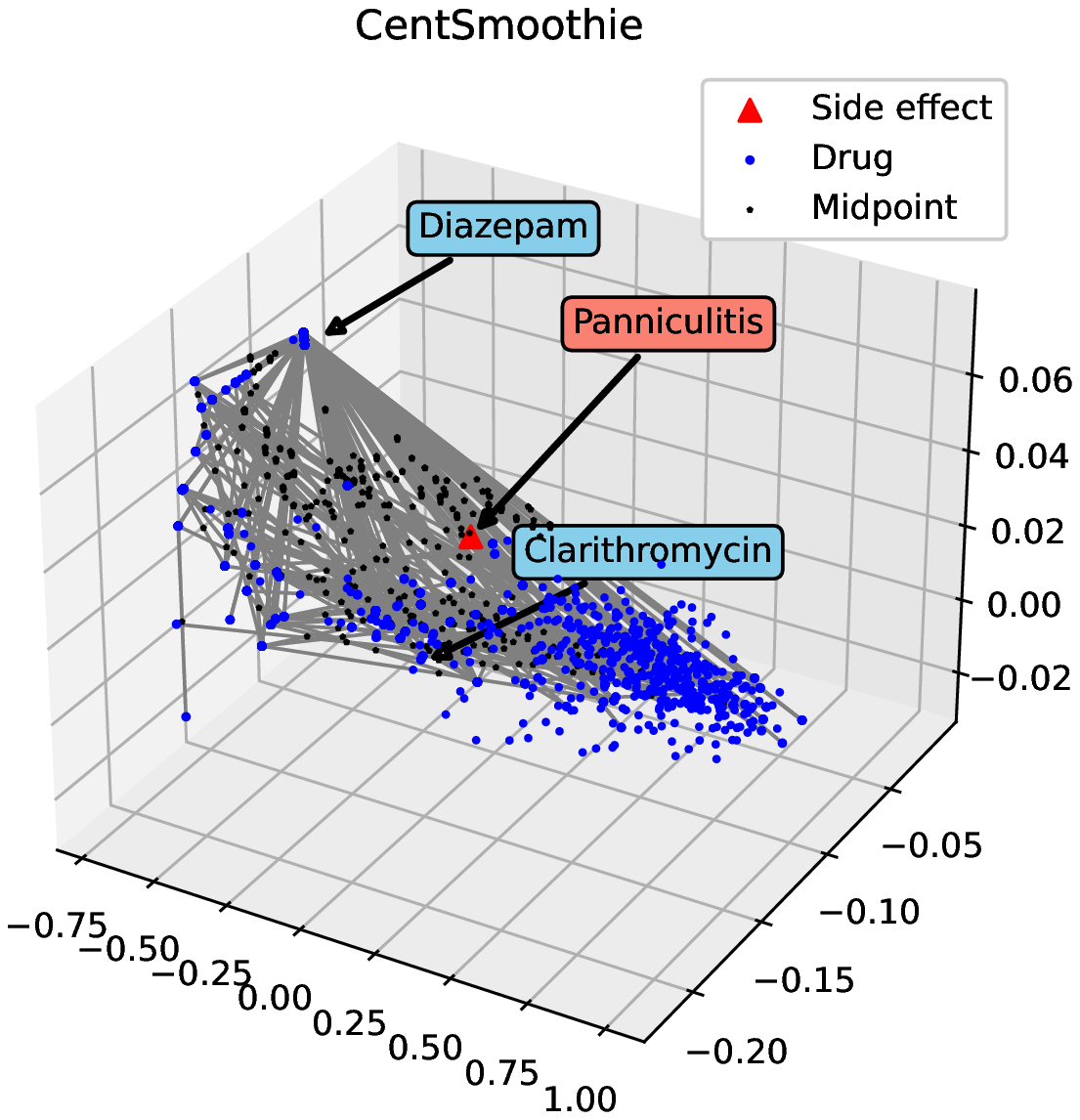}\label{fig:newV1}}

\subfloat[][]{\includegraphics[width=0.3\textwidth]{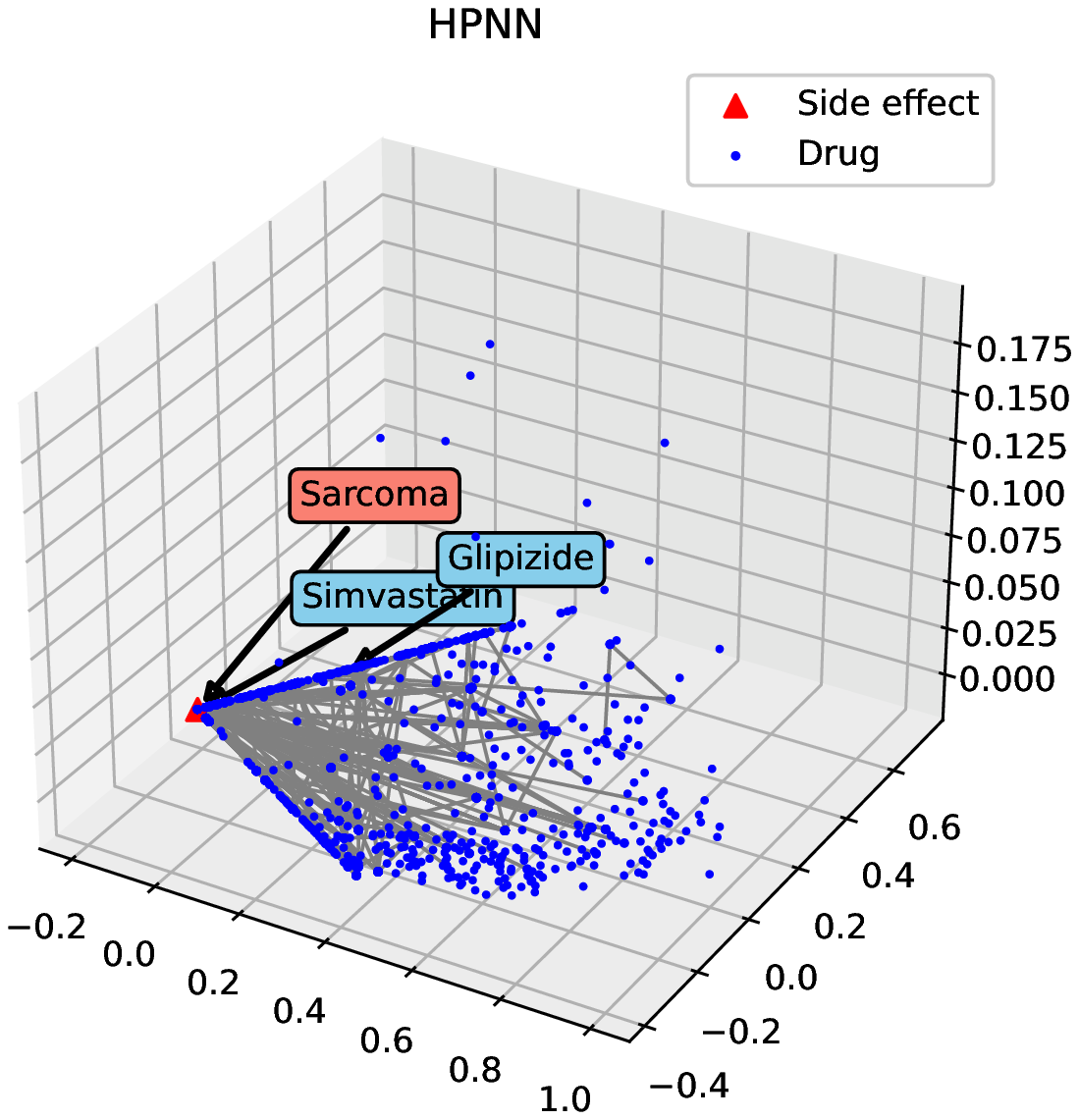}\label{fig:oldV2}}
\subfloat[][]{\includegraphics[width=0.3\textwidth]{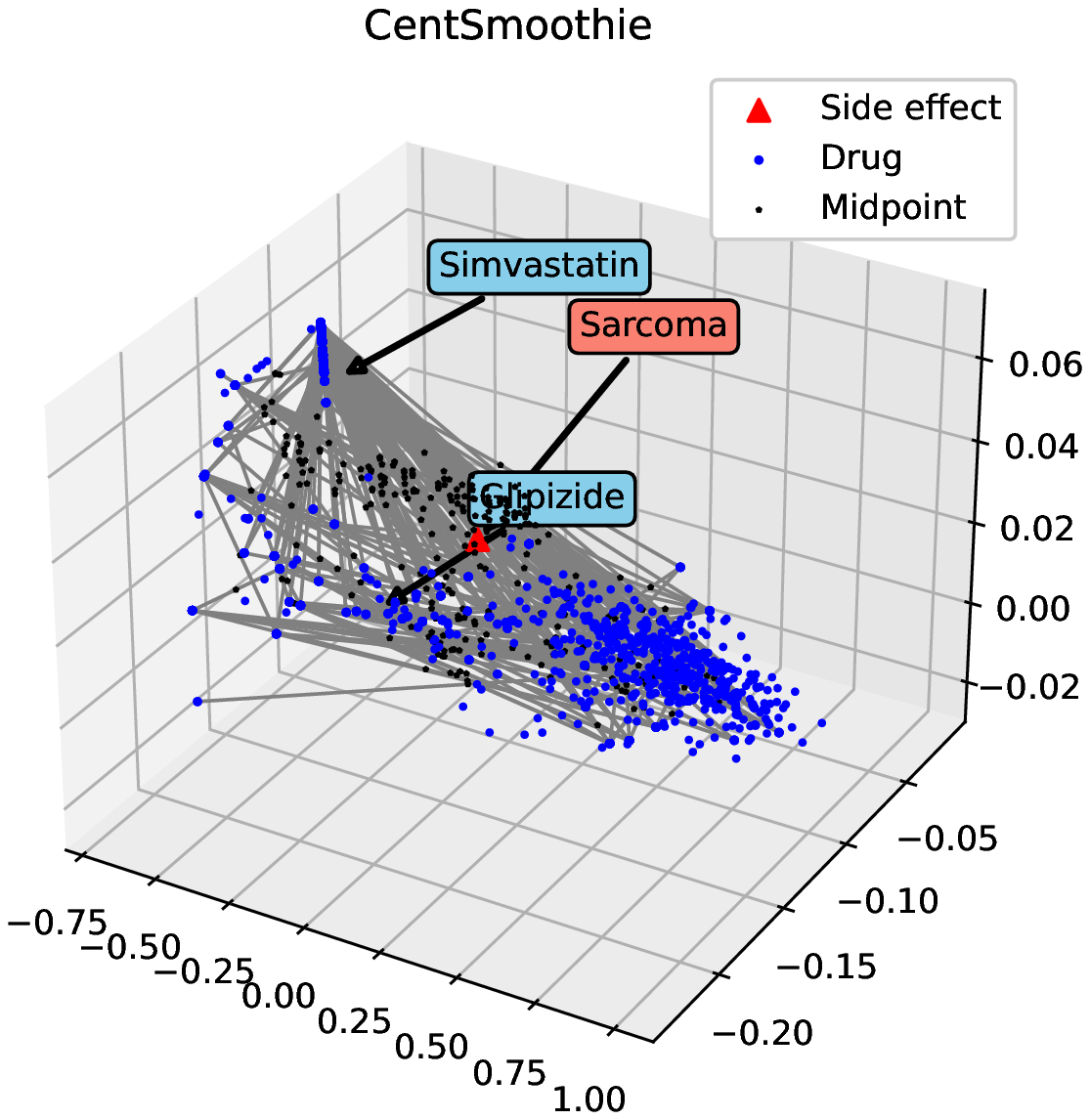}\label{fig:newV2}}

\subfloat[][]{\includegraphics[width=0.3\textwidth]{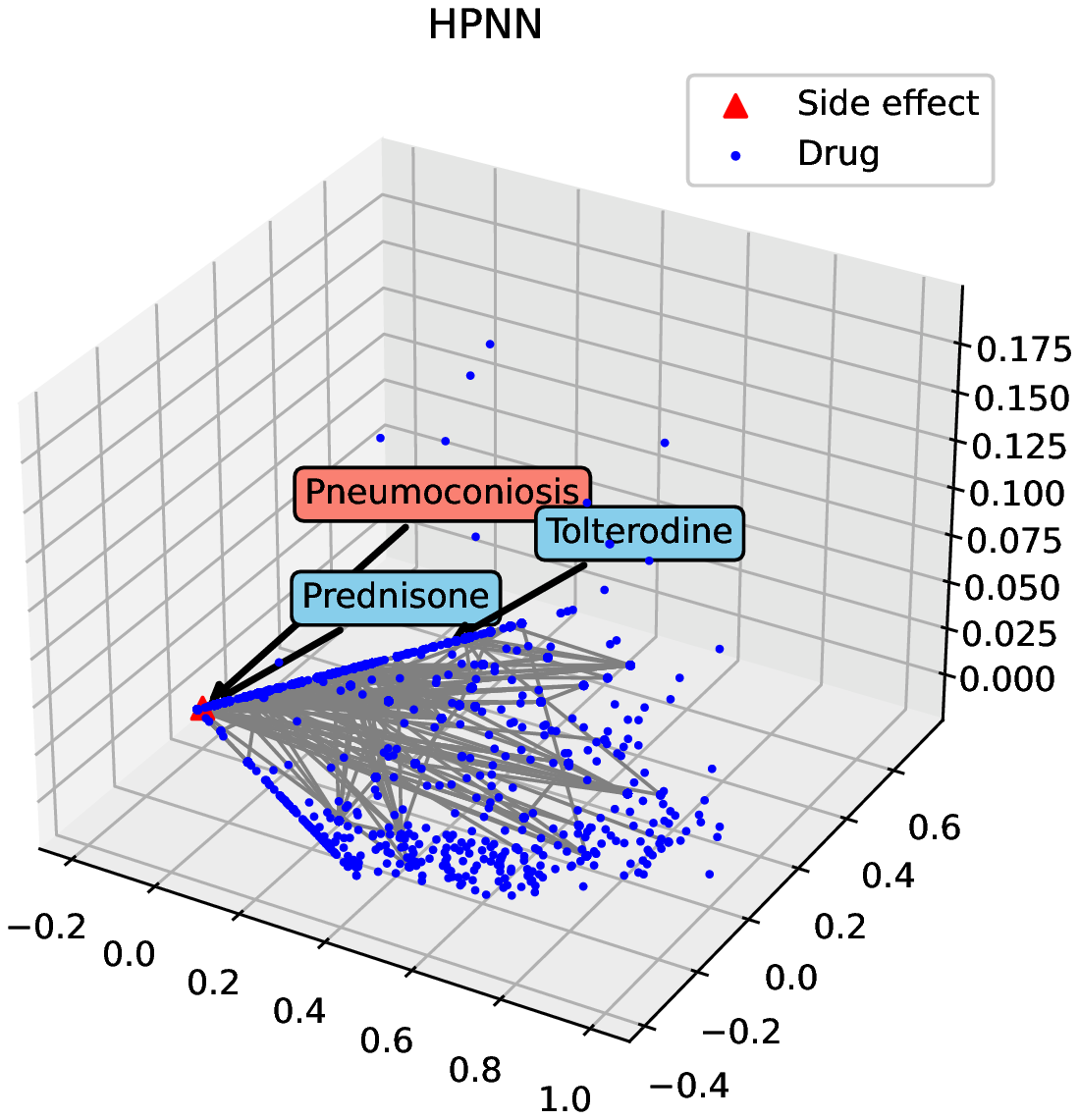}\label{fig:oldV3}}
\subfloat[][]{\includegraphics[width=0.3\textwidth]{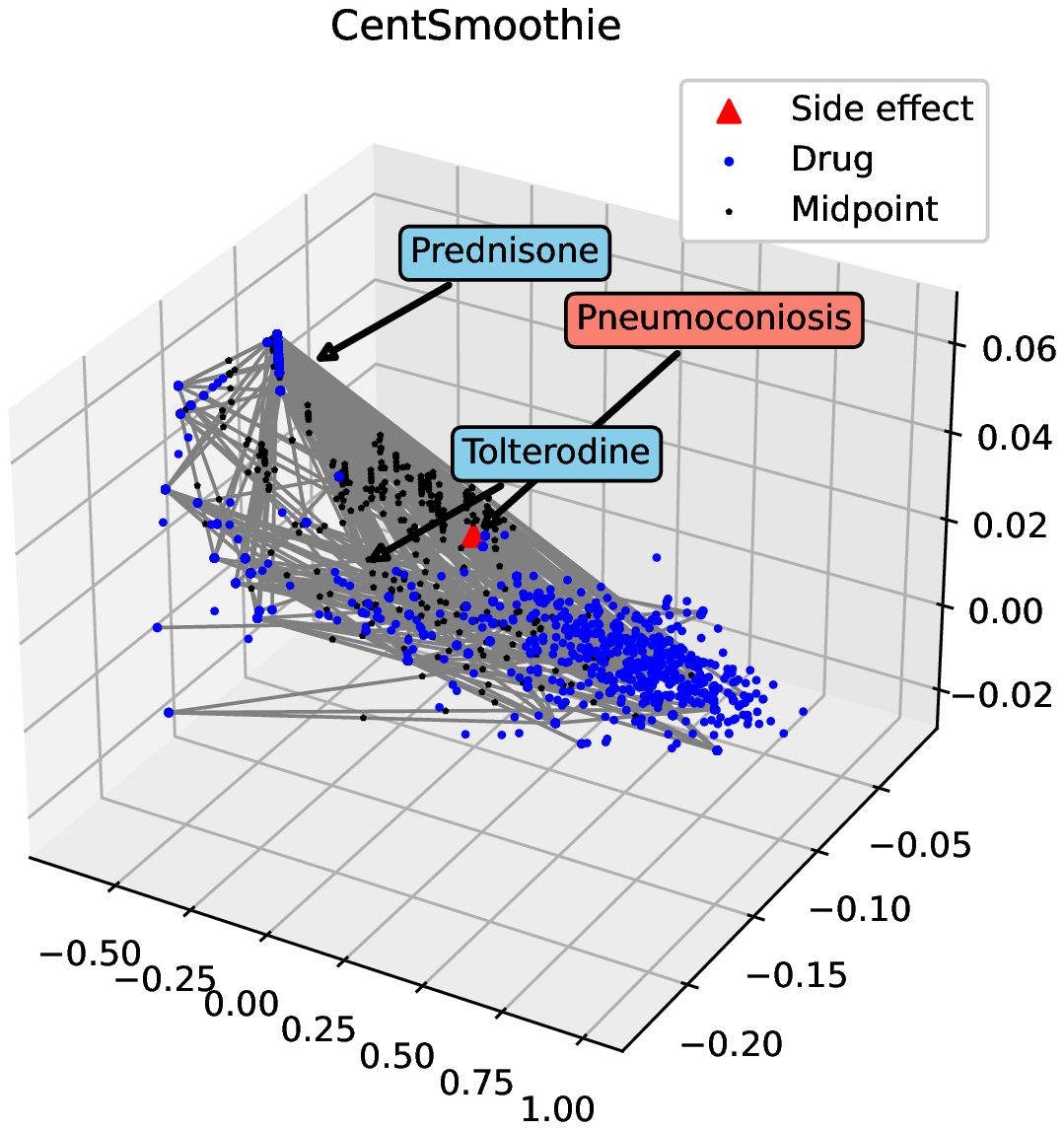}\label{fig:newV3}}

\subfloat[][]{\includegraphics[width=0.3\textwidth]{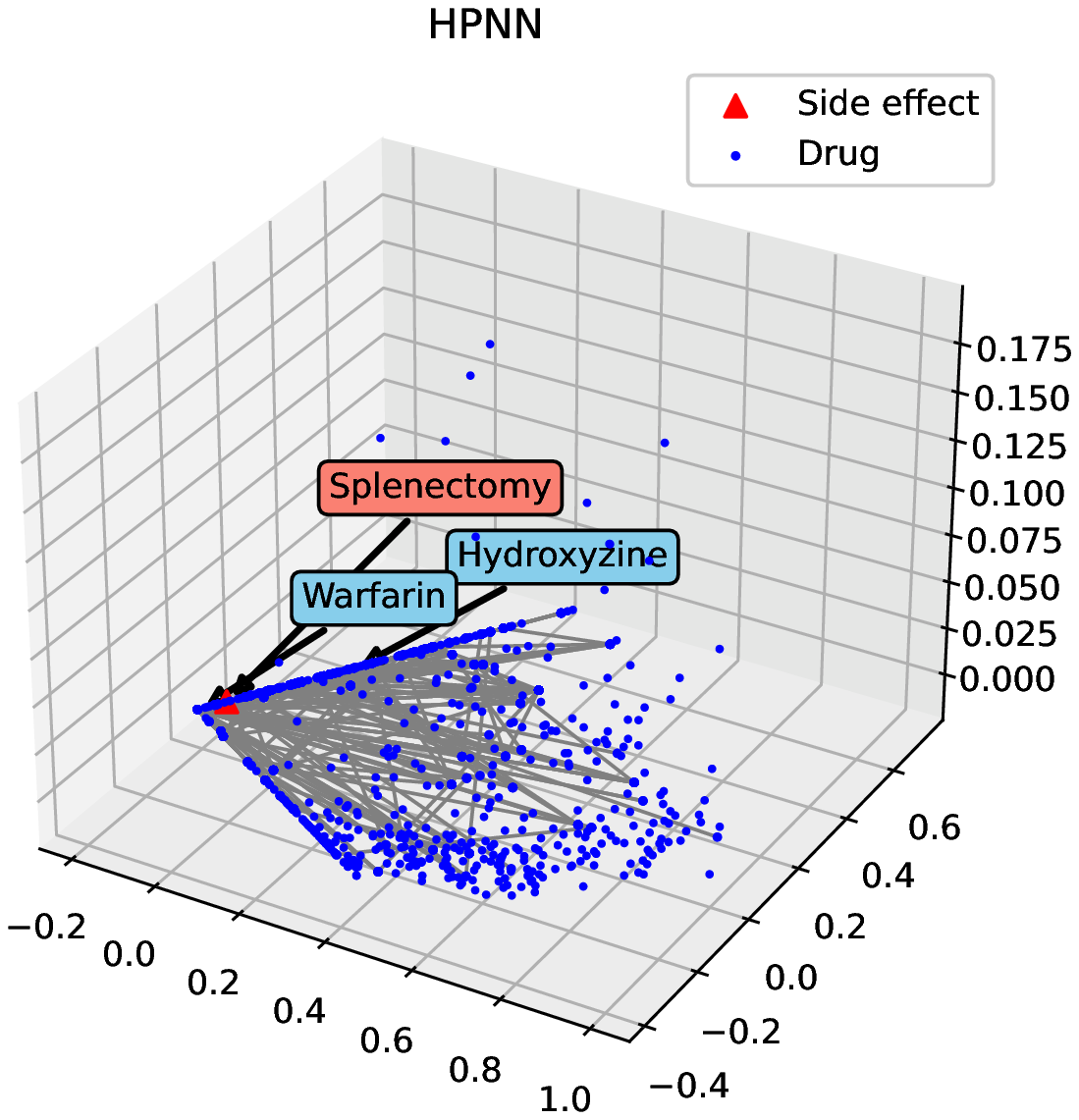}\label{fig:oldV4}}
\subfloat[][]{\includegraphics[width=0.3\textwidth]{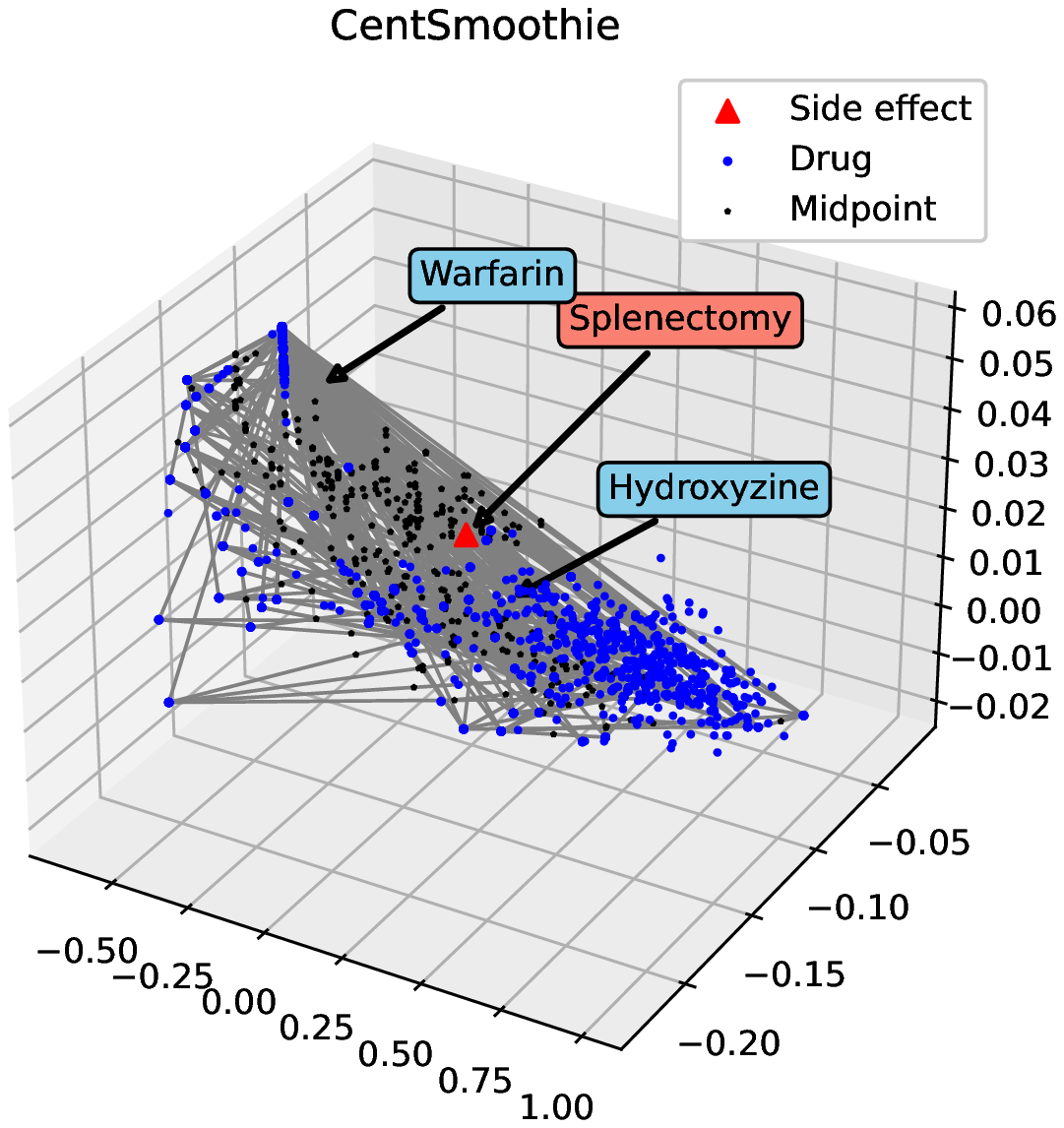}\label{fig:newV4}}

\caption{Visualization of representations of drugs and side effects ((a-b) Panniculitis, (c-d) Sarcoma, (e-f) Pneumoconiosis, and (g-h) Splenectomy) learnt from HPNN and CentSmoothie trained with TWOSIDES. In CentSmoothie, the representation of a side effect tends to close to the mean of all midpoints of drug pairs causing the side effect. In HPNN, the representation of the side effect is hard to distinguish from the drugs. }
\label{fig:visual}
\end{figure*}

\section*{Case studies for predicting unknown drug pairs on infrequent side effects}
\label{sub:case}

\begin{table*}[htbp]
    \centering
    \begin{tabular}{lrrrrrrrrrrrrr}
    \hline
    \multirow{2}{*}{Side effect} & \multirow{2}{*}{Drug pair} & \multicolumn{3}{c}{Rank (Score)} & \multirow{2}{*}{Literature} \\ \cline{3-5}
    & & $\mathrm{CentSmoothie}$ & $\mathrm{HPNN}$ & Decagon & \\ \hline 
 \multirow{3}{*}{Panniculitis} & Ranitidine, Pioglitazone & 1(0.94) & 10(0.53) & - & \checkmark \\ \cline{2-6}
& Diazepam, Clarithromycin & 2(0.94) & 7(0.57) & 139(0.27) & \checkmark \\ \cline{2 - 6}
& Folic Acid, Metoclopramide & 3(0.89) & 12(0.50) & 62(0.40) & - \\ \cline{2 - 6}
& Fexofenadine, Furosemide & 4(0.88) & 6(0.58) & 34(0.47) & \checkmark \\ \cline{2 - 6}
& Metronidazole, Salbutamol & 5(0.87) & 5(0.59) & 1(0.61) & \checkmark \\ \cline{2 - 6}
& Zolpidem, Warfarin & 6(0.85) & 1(0.66) & 91(0.34) & \checkmark \\ \cline{2 - 6}
& Salbutamol, Warfarin & 7(0.85) & 2(0.66) & - & \checkmark \\ \cline{2 - 6}
& Sertraline, Hydrochlorothiazide & 8(0.85) & 4(0.62) & 130(0.29) & - \\ \cline{2 - 6}
& Warfarin, Tolterodine & 9(0.84) & 17(0.45) & - & \checkmark \\ \cline{2 - 6}
& Acetaminophen, Amoxicillin & 10(0.82) & 13(0.48) & 61(0.40) & \checkmark \\ \hline
\multirow{3}{*}{Sarcoma} & Carvedilol, Ramipril & 1(0.80) & 2(0.61) & - & \checkmark \\ \cline{2-6}
& Simvastatin, Glipizide & 2(0.80) & 10(0.49) & 21(0.50) & \checkmark \\ \cline{2 - 6}
& Ibuprofen, Mirtazapine & 3(0.80) & 13(0.44) & 45(0.46) & \checkmark \\ \cline{2 - 6}
& Lactulose, Simvastatin & 4(0.79) & 11(0.46) & 55(0.42) & \checkmark \\ \cline{2 - 6}
& Zolpidem, Fluticasone & 5(0.78) & 1(0.66) & 89(0.35) & - \\ \cline{2 - 6}
& Prednisolone, Acetaminophen & 6(0.78) & 5(0.57) & - & \checkmark \\ \cline{2 - 6}
& Lisinopril, Nystatin & 7(0.77) & 17(0.44) & 25(0.48) & \checkmark \\ \cline{2 - 6}
& Ibuprofen, Fluticasone & 8(0.77) & 3(0.61) & 31(0.48) & \checkmark \\ \cline{2 - 6}
& Acetylsalicylic acid, Alendronic acid & 9(0.76) & 6(0.57) & - & \checkmark \\ \cline{2 - 6}
& Fluticasone, Famotidine & 10(0.74) & 4(0.60) & - & \checkmark \\ \hline
\multirow{3}{*}{Pneumoconiosis} & Prednisone, Tolterodine & 1(0.91) & 2(0.58) & 2(0.62) & \checkmark \\ \cline{2-6}
& Celecoxib, Diltiazem & 2(0.88) & 1(0.60) & - & - \\ \cline{2 - 6}
& Atorvastatin, Fenofibrate & 3(0.80) & 7(0.46) & 170(0.23) & \checkmark \\ \cline{2 - 6}
& Rosuvastatin, Acetaminophen & 4(0.78) & 3(0.50) & - & \checkmark \\ \cline{2 - 6}
& Losartan, Carisoprodol & 5(0.77) & 10(0.43) & 187(0.21) & - \\ \cline{2 - 6}
& Oxycodone, Zoledronic acid & 6(0.68) & 5(0.49) & 18(0.51) & - \\ \cline{2 - 6}
& Gabapentin, Diclofenac & 7(0.68) & 4(0.50) & 20(0.50) & \checkmark \\ \cline{2 - 6}
& Risedronate, Metoclopramide & 8(0.68) & 11(0.43) & 49(0.43) & - \\ \cline{2 - 6}
& Rofecoxib, Pamidronate & 9(0.66) & 9(0.44) & - & \checkmark \\ \cline{2 - 6}
& Tamsulosin, Ofloxacin & 10(0.65) & 6(0.49) & 4(0.61) & \checkmark \\ \hline
\multirow{3}{*}{Splenectomy} & Doxycycline, Alendronic acid & 1(0.84) & 5(0.59) & - & \checkmark \\ \cline{2-6}
& Hydroxyzine, Warfarin & 2(0.83) & 7(0.54) & 109(0.38) & - \\ \cline{2 - 6}
& Paroxetine, Pamidronate & 3(0.80) & 4(0.62) & 87(0.40) & \checkmark \\ \cline{2 - 6}
& Oxycodone, Venlafaxine & 4(0.80) & 3(0.63) & - & \checkmark \\ \cline{2 - 6}
& Lorazepam, Acetaminophen & 5(0.76) & 1(0.75) & - & \checkmark \\ \cline{2 - 6}
& Zolpidem, Lansoprazole & 6(0.76) & 2(0.74) & - & \checkmark \\ \cline{2 - 6}
& Hydroxyzine, Bupropion & 7(0.73) & 12(0.47) & 3(0.61) & \checkmark \\ \cline{2 - 6}
& Paroxetine, Niacin & 8(0.71) & 16(0.43) & 153(0.30) & - \\ \cline{2 - 6}
& Simvastatin, Doxazosin & 9(0.71) & 15(0.44) & - & \checkmark \\ \cline{2 - 6}
& Enalapril, Cephalexin & 10(0.68) & 10(0.49) & 58(0.50) & \checkmark \\ \hline
    
    \end{tabular}
    \caption{Predictions of unknown drug pairs for a side effect, top-ranked by CentSmoothie (trained with TWOSIDES) with prediction scores and evidences from the literature.}
    \label{tb:casestudies}
\end{table*}

We showed sampled results obtained by $\mathrm{CentSmoothie}$ trained with the largest dataset (TWOSIDES), for predicting unknown drug pairs of each side effect, where the drug pairs with the side effect shown here are not in the current drug-drug interaction data \cite{tatonetti2012data}. Our focus was on infrequent side effects, which were thought to be harder to predict.
Also, we confirmed the biological validity of the predicted drug pairs by finding relevant biomedical articles through searching the biological literature using keywords of the predicted drug pair and the side effect.

Table \ref{tb:casestudies} shows the four side effects (selected from the top 5\% infrequent side effects), and for each side effect, ten unknown pairs with the highest prediction scores by $\mathrm{CentSmoothie}$.
Also for each of the ten drug pairs, the score obtained by HPNN (also Decagon) and the rank according to the score are shown if they were in top 200 predictions. The last column showed the article relevant to each predicted drug pair.
For 31 of the 40 predictions, we could find evidence (biomedical articles) by literature survey, implying the prominent of the findings by $\mathrm{CentSmoothie}$.
Comparing with the ranks (top ten) by $\mathrm{CentSmoothie}$, those by HPNN were larger. Meanwhile, those ranks by Decagon were very large, where some rankes were out of top 200, meaning that CentSmoothie and Decagon have different prediction preferences.

Taking a closer look, for example, for sarcoma, the highest score was achieved by the pair of Carvedilol and Ramipril, where this pair was not in the training drug-drug interaction data \cite{tatonetti2012data}, while this new interaction could be predicted by $\mathrm{CentSmoothie}$ and validated by \cite{gujral2018effect}. These results demonstrated that prediction by $\mathrm{CentSmoothie}$ for unknown drug pairs could be used for further clinical verification and so $\mathrm{CentSmoothie}$ itself would be a highly useful model.

 \section*{Visualizing representations}

\subsection*{Side effects and drug pairs}

We visualized the representations of drugs and side effects learnt by $\mathrm{CentSmoothie}$ and $\mathrm{HPNN}$ using TWOSIDES dataset to examine the difference between the central-smoothing assumption and the traditional smoothing assumption. We used the same four side effects as those we showed in Section \ref{sub:case}.

Fig. \ref{fig:visual} shows the visualization obtained by applying principal component analysis (PCA) to the resultant representation by each of the two methods, where for each side effect, drugs (blue dots) and the side effect (red triangle) are shown in the three-dimensional (3D) space.
(For $\mathrm{CentSmoothie}$, the representations on the subspace corresponding to the side effect were fed into PCA). 
We drew (gray) lines for drug pairs with side effects. For $\mathrm{CentSmoothie}$, we further showed the midpoint of each drug pair (with a side effect) by a black dot, to see if the midpoint is close to the representation of the side effect. We could easily see that for each side effect, the representations of side effects tended to be located around at the mean point among all midpoints (black dots). 
%This suggested that if a midpoint is close to the side effect, then the corresponding drug pair likely causes the side effect. 
However, for HPNN, it was difficult to interpret the representations (of side effects) learnt by HPNN among the representations of drugs.
Also by using these visualizations, we could easily understand how each pair of drugs and the side effect are positioned in the space. Particularly, by checking if the side effect is located nearby the midpoint of the corresponding drug pair, we can guess that the side effect might be caused. 

\subsection*{Side effects relationships}

\begin{figure}
    \centering
    \includegraphics[width=0.5\textwidth]{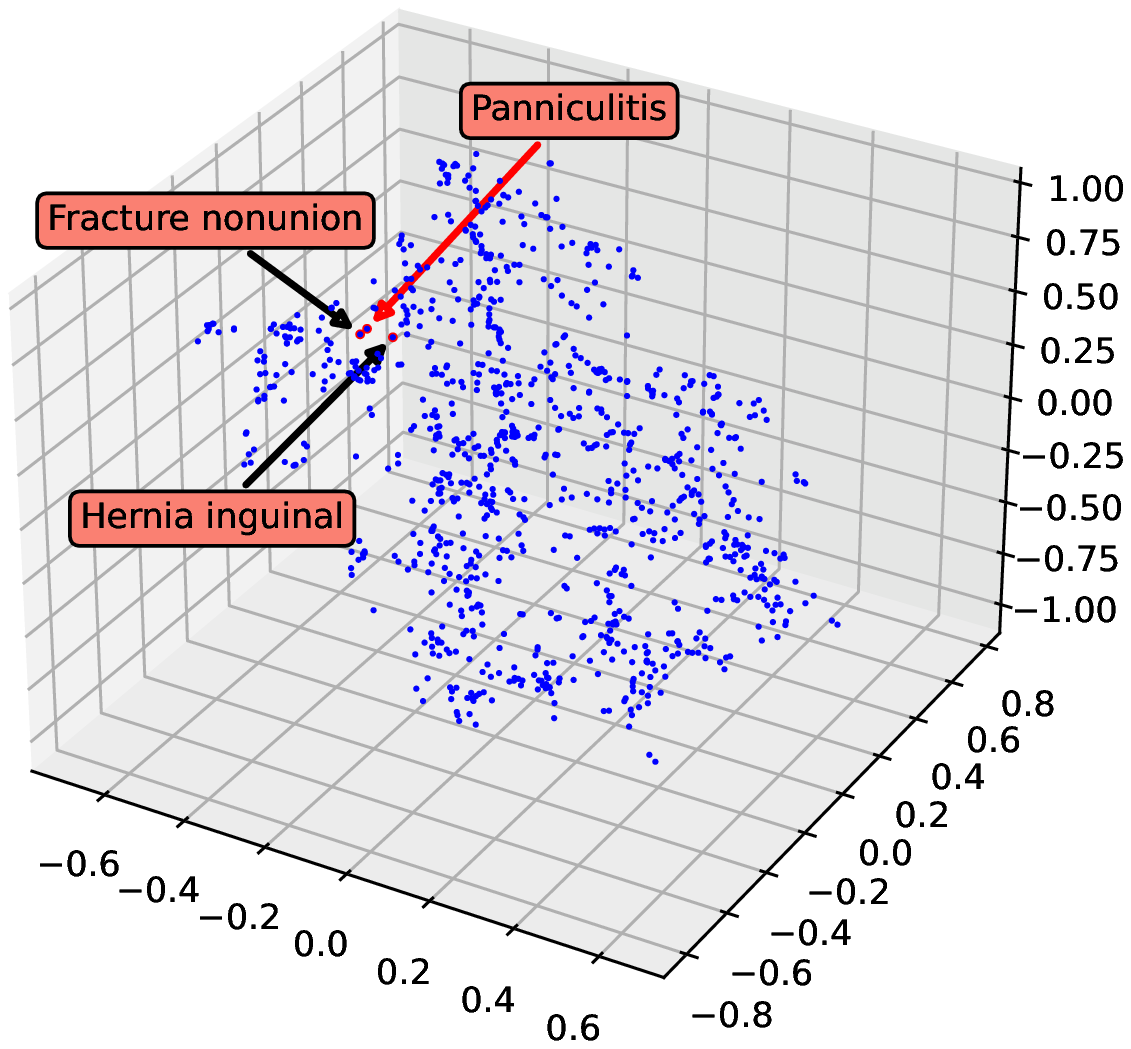}
    \caption{Visualization of side effect representations.}
    \label{fig:serel}
\end{figure}

We visualized the representations of all side effects learnt by $\mathrm{CentSmoothie}$ on TWOSIDES dataset to see the relationships of side effects. Fig. \ref{fig:serel} shows the visualization of side effects in a three-dimensional space. We could see that side effects are grouped into some small clusters. We highlighted an infrequent side effect: Panniculitis and two of its nearest neighbors: Fracture nonunion and Hemia inguinal. Furthermore we could find the evidence for the occurrence of Panniculitis with Fracture nonunion and Hemia inguinal \cite{ogden1960panniculitis,stieger2015extracorporeal}.

\end{document}